%% file: X2N.tex
\documentclass[letterpaper, 10 pt, conference]{ieeeconf} 

\IEEEoverridecommandlockouts                              

\usepackage{graphics} 
\usepackage{epsfig} 
\usepackage{times} 
\usepackage{amsmath} 
\usepackage{amssymb}  
\usepackage{amsfonts}
\usepackage{tabularx}
\usepackage[sort,compress]{cite}
\usepackage[colorlinks,linkcolor=black,anchorcolor=black,citecolor=black,urlcolor=black]{hyperref}
\usepackage[table]{xcolor}
\usepackage{booktabs}
\usepackage{multirow}
\usepackage{graphicx}
\usepackage{subfigure}
\usepackage{float}
\usepackage{makecell}
\usepackage{booktabs}
\usepackage{etoolbox}
\usepackage{bbding}
\usepackage{makecell}
\usepackage{caption}
\usepackage{threeparttable}
\makeatletter
\patchcmd{\@makecaption}
  {\scshape}
  {}
  {}
  {}
\makeatletter
\patchcmd{\@makecaption}
  {\\}
  {.\ }
  {}
  {}
\makeatother

\graphicspath{{../img/}}

\title{\LARGE \bf
X2-N: A Transformable Wheel-legged Humanoid Robot \\ with Dual-mode Locomotion and Manipulation
}
\author{Yan Ning$^{1}$, Xingzhou Chen$^{1}$, Delong Li$^{2}$, Hao Zhang$^{2}$, Hanfu Gai$^{2}$, Jinlei Zhao$^{2}$, \\Tongyuan Li$^{2}$, Cheng Zhang$^{2}$, Zhihui Peng$^{2}$ and Ling Shi$^{1}$
\thanks{$^{1}$Department of Electronic and Computer Engineering, The Hong Kong University of Science and Technology, Hong Kong, China.}
\thanks{$^{2}$X-lab Department, Agibot Innovation Company, Shanghai, China.}
\thanks{Email: {\tt\footnotesize
\{yningaa, xchenfk\}@connect.ust.hk, \newline
\hspace*{4em}lidelong@agibot.com, ahaobupt@gmail.com,\newline
\hspace*{4em}gaihanfu@live.com, in\_barrel@outlook.com,\newline
\hspace*{4em}\{zhangcheng, pengzhihui\}@zhiyuan-robot.com,\newline
\hspace*{4em}eesling@ust.hk
}}
}
\begin{document}


\makeatletter
\let\@oldmaketitle\@maketitle
\renewcommand{\@maketitle}{\@oldmaketitle

  \bigskip}
\makeatother

\maketitle

\begin{abstract}
Wheel-legged robots combine the efficiency of wheeled locomotion with the versatility of legged systems, enabling rapid traversal over both continuous and discrete terrains. However, conventional designs typically employ fixed wheels as feet and limited degrees of freedom (DoFs) at the hips, resulting in reduced stability and mobility during legged locomotion compared to humanoids with flat feet. In addition, most existing platforms lack a full upper body with arms, which limits their ability to perform dexterous manipulation tasks.

In this letter, we present X2-N, a high-DoF transformable robot with dual-mode locomotion and manipulation. X2-N can operate in both humanoid and wheel-legged forms and transform seamlessly between them through joint reconfiguration. We further propose a reinforcement learning (RL)-based whole-body control framework tailored to this morphology, enabling control across hybrid locomotion, transformation, and manipulation. We validate X2-N in a range of challenging locomotion and manipulation tasks, including dynamic skating-like motion, stair climbing and package delivery. Results demonstrate high locomotion efficiency, strong terrain adaptability, and stable loco-manipulation performance of X2-N, highlighting its potential for real-world deployment.

\end{abstract}
\section{Introduction}
\label{sec:intro}
In recent years, agile legged robots with high degrees of freedom (DoFs) have achieved significant progress in both research and industry. Among them, humanoid and wheel-legged robots have emerged as two prominent platforms.

Humanoid robots \cite{BostonDynamics2024NewAtlas, Unitree2025G1, Figure2026Helix02LivingRoom} exhibit impressive capabilities in robust dynamic locomotion and dexterous manipulation through a high-DoF full-body design. Their bipedal structure with flat feet also provides robust standing with minimal ground occupancy, thereby benefiting static manipulation. However, during movement, legged locomotion involves frequent impacts to the body and compromises operational stability. Moreover, on continuous and flat terrains typical in industrial or civil scenarios, legged locomotion is generally less energy efficient, slower, and less stable compared to wheel-legged systems.

Wheel-legged robots integrate wheels into legged systems to improve locomotion efficiency and terrain adaptability \cite{Rolling, LimXDynamics2024TRON1, diablo}. They can achieve stable operation during smooth wheeled locomotion, but suffer from unsteady point-contact constraints during legged locomotion. Moreover, most systems are designed with limited-DoF legs and only lower bodies, restricting their capabilities in manipulation compared to humanoids. Therefore, it remains challenging for existing robotic platforms to simultaneously achieve efficient, agile locomotion together with stable, dexterous manipulation in a unified form for practical applications.

\begin{figure}[t]
    \centering
    \vspace{0.15cm}
    \includegraphics[width=0.93\linewidth]{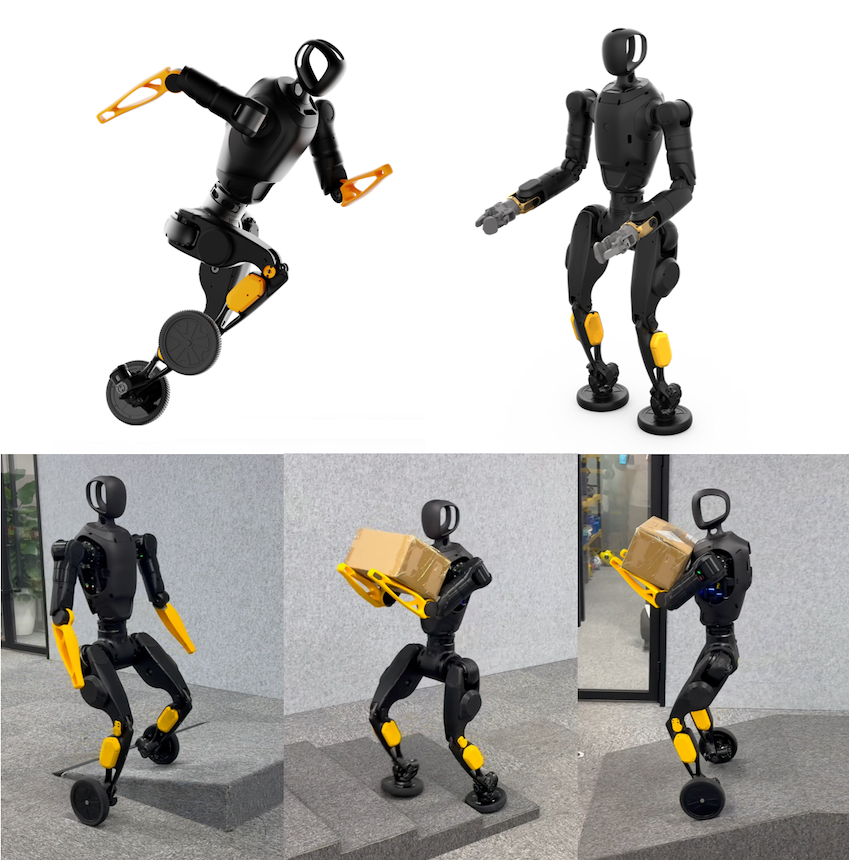}
    \caption{Illustration of X2-N in dual locomotion modes with 4/7-DoF arms, operating on stairs and slopes.}
    \label{fig:overview}
    \vspace{-0.45cm}
\end{figure}

In this paper, we present X2-N, a high-DoF transformable wheel-legged humanoid robot with dual working modes. The robot can operate in either wheel-contact or foot-contact configurations for agile and efficient locomotion, and can interchange between 4-DoF and 7-DoF arms for manipulation. X2-N can transition seamlessly between these modes for the most suitable, efficient and stable operation form with a compact size and weight.

To control this dual-mode system, we develop a hierarchical control framework tailored to the proposed morphology. It combines a reinforcement learning (RL)-based controller with a model-based whole-body controller for hybrid locomotion, mode transformation, and loco-manipulation. We validate the effectiveness and robustness of X2-N through simulation and hardware experiments. 

Our \textbf{contributions} are summarized as follows:

(1) We present X2-N, a novel transformable robot platform that integrates features of both wheel-legged and humanoid robots. The system is built upon customized hardware and actuators, enabling both efficient locomotion and dexterous manipulation within a unified design.

(2) We propose a new lower-limb joint configuration and mechanism that support both wheel- and foot-contact modes without introducing redundant DoFs for transformation and steering. The legs of X2-N preserve sufficient workspace and maneuverability comparable to existing humanoid legs.

(3) We develop a hybrid hierarchical control framework for the proposed dual-mode transformable morphology. The framework enables X2-N to switch between specific RL-based agents with a shared whole-body model-based controller for the most suitable operation form.

(4) We will open-source parts of the robot design, hardware and software interfaces, and control framework to enable researchers to customize and further develop on this transformable robot platform.

\section{Related Work}

\label{sec:related_works}

\subsection{Humanoid Platforms}

Humanoid robots have achieved remarkable progress in dynamic locomotion and 
dexterous manipulation through high-DoF full-body structures and articulated 
limbs. Their anthropomorphic morphology provides large workspace, versatile 
contact capability, and effective whole-body interaction with human-oriented 
environments. Representative platforms such as Atlas~\cite{BostonDynamics2024NewAtlas}, 
G1~\cite{Unitree2025G1}, and Helix~\cite{Figure2026Helix02LivingRoom} 
demonstrate agile locomotion, dynamic balancing, and generalized operation 
for practical applications.

However, existing humanoid robots are generally optimized for a fixed 
morphology, which inherently introduces a trade-off between locomotion 
efficiency and operational capability. Although legged locomotion provides 
excellent terrain adaptability and whole-body interaction ability, repeated 
foot-ground contacts introduce additional energy consumption and motion 
complexity, especially during long-distance traversal. Therefore, achieving 
both efficient mobility and dexterous operation remains a fundamental 
challenge for humanoid robot design.

\begin{figure}[t]
    \centering
    \vspace{0.15cm}
    \includegraphics[width=0.9\linewidth]{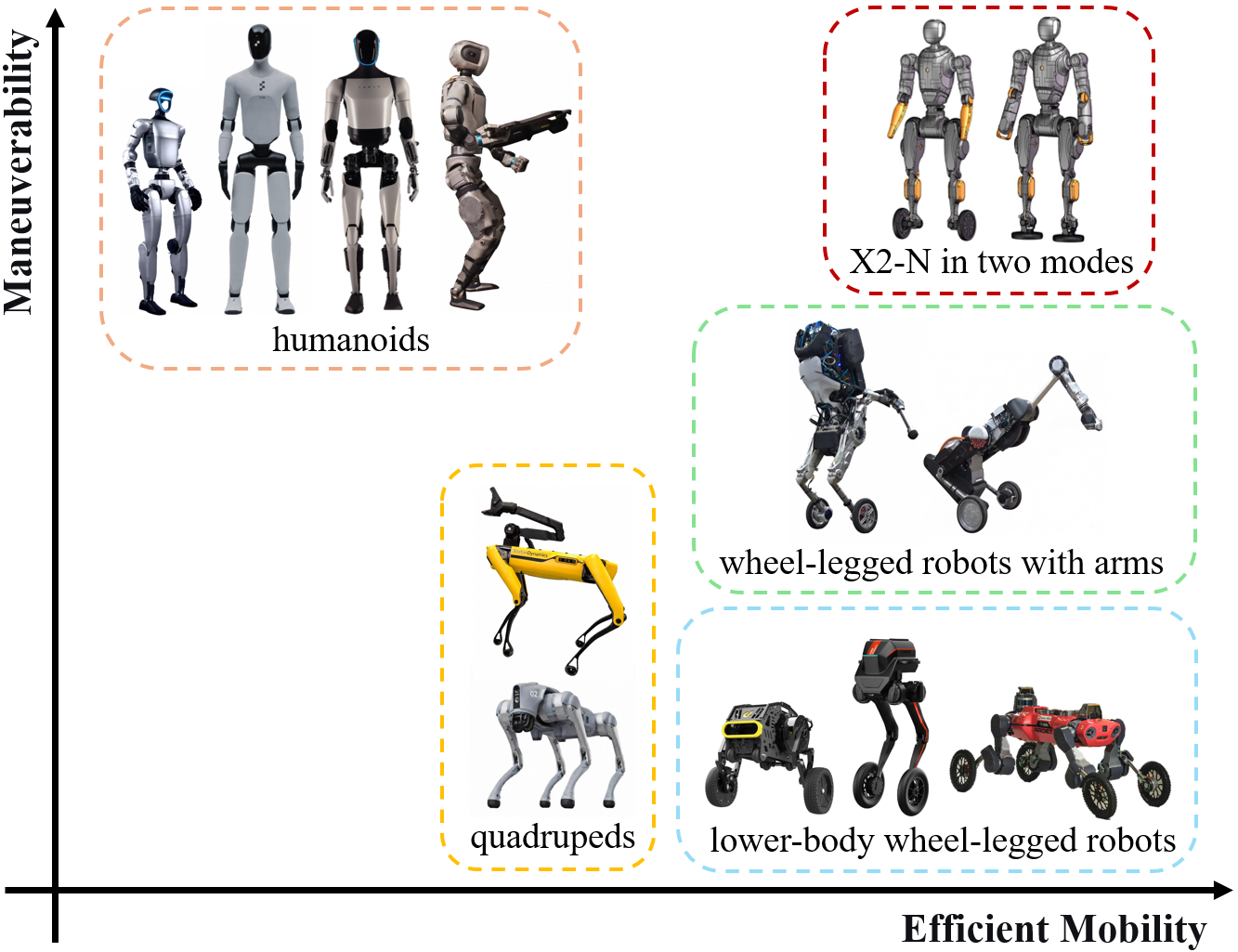}
    \caption{Overview of different classic legged robot platforms.}
    \label{fig:rela}
    \vspace{-0.4cm}
\end{figure}

\subsection{Wheel-Legged Platforms}

Wheel-legged robots combine the terrain adaptability of legged systems with the efficiency of wheeled locomotion. Representative quadrupedal platforms such as ANYmal~\cite{learningonANymal} and B2-W~\cite{unitree_b2w_2026} demonstrate robust mobility across challenging terrains, while compact bipedal platforms including Tron~1~\cite{LimXDynamics2024TRON1} and  DIABLO~\cite{diablo} achieve agile locomotion in constrained environments. Compared with conventional humanoids, these platforms significantly improve locomotion efficiency on continuous and structured terrains.

Despite their mobility advantages, most existing wheel-legged robots are primarily designed with locomotion efficiency as the first priority. Their simplified limb structures and lower-body-focused configurations reduce mechanical complexity but limit dexterous manipulation and whole-body interaction capabilities. Although several platforms introduce lightweight manipulators for loco-manipulation~\cite{cheng2025rambo, badminton}, their limited DoF and non-humanoid configurations still restrict versatile operation compared with humanoid systems. Furthermore, wheeled locomotion usually provides limited support configurations for complex manipulation tasks.

Similar to humanoid robots, current wheel-legged platforms generally adopt fixed morphologies. A robotic platform that can combine the advantages of both morphologies remains largely unexplored.

\subsection{Control of High-DoF Legged Robot}

Whole-body control of high-DoF robots has been dominated by model-based optimization methods, such as inverse dynamics, quadratic programming, and model predictive control~\cite{mit_human3}. These methods provide accurate motion tracking and force regulation but rely heavily on model accuracy and computational efficiency, limiting their robustness under highly dynamic and uncertain contact conditions.

Reinforcement learning has recently enabled agile locomotion and loco-manipulation with improved robustness against modeling uncertainties ~\cite{RLCassie, zhang2025falcon}. However, existing learning-based methods mainly focus on dynamic locomotion and generally lack the precise manipulation capability of model-based controllers.

Hybrid control frameworks have recently combined learning-based locomotion policies with model-based whole-body controllers, leveraging the advantages of both approaches~\cite{learningonANymal, badminton}. However, existing control frameworks mainly assume fixed robot morphologies and predefined contact configurations. For robots with multiple operating mode, the controller must have additionally coordination across different configurations, which remains underdeveloped.

Overall, existing humanoid and wheel-legged robots represent two distinct design paradigms, where humanoids prioritize dexterous whole-body interaction while wheel-legged systems emphasize efficient mobility. However, these advantages are difficult to simultaneously achieve within a fixed morphology. A transformable robotic platform that can dynamically switch between efficient wheeled locomotion and humanoid-level operation remains unexplored. Moreover, such a system requires a control framework capable of coordinating locomotion, transformation, and manipulation across different morphologies. These challenges motivate the development of X2-N presented in this work.

\section{System Design}
\begin{figure}[t]
    \centering
    \vspace{0.15cm}
    \includegraphics[width=0.9\linewidth]{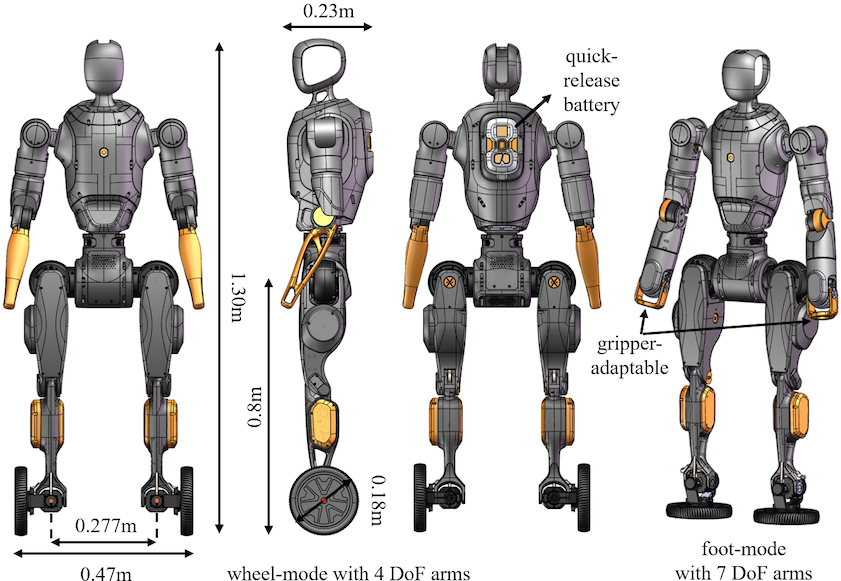}
    \caption{Illustration of X2-N's specifications.}
    \label{fig:design}
    \vspace{-0.5cm}
\end{figure}

\subsection{Overview}
X2-N is a fully electric and mid-sized wheel-legged humanoid robot with both wheel-legged and foot-legged modes. It weighs approximately 30.7 kg and stands 1.3 m tall, as shown in Fig. \ref{fig:design}. The system consists of a torso, a waist, two arms, and two legs, primarily constructed from aluminum. 

The standard configuration of X2-N has 21 DoFs in foot-legged mode and 17 DoFs in wheel-legged mode to accommodate the distinct requirements of the two locomotion modes. Through a unified mechanical design, the system supports both transformation and locomotion without introducing additional actuators, thereby avoiding increased system weight and complexity. The upper body adopts a modular design with interchangeable manipulators, supporting both lightweight operation and dexterous manipulation. Detailed designs of the lower limbs, upper body, and actuation system are presented in the following sections.

\subsection{A Two-mode Transformable Leg Design}
Designing a unified robot platform that integrates humanoid and wheel-legged configurations is challenging due to their mismatched kinematic structures, particularly in degrees of freedom (DoFs) and contact constraints. Moreover, enabling transformation between two modes typically requires additional dedicated actuation. To address this issue, we propose a transformable lower-limb architecture based on a joint reuse principle, to support both locomotion and transformation within a unified design. 

\begin{figure}[t]
    \centering
    \vspace{0.15cm}
    \includegraphics[width=0.95\linewidth]{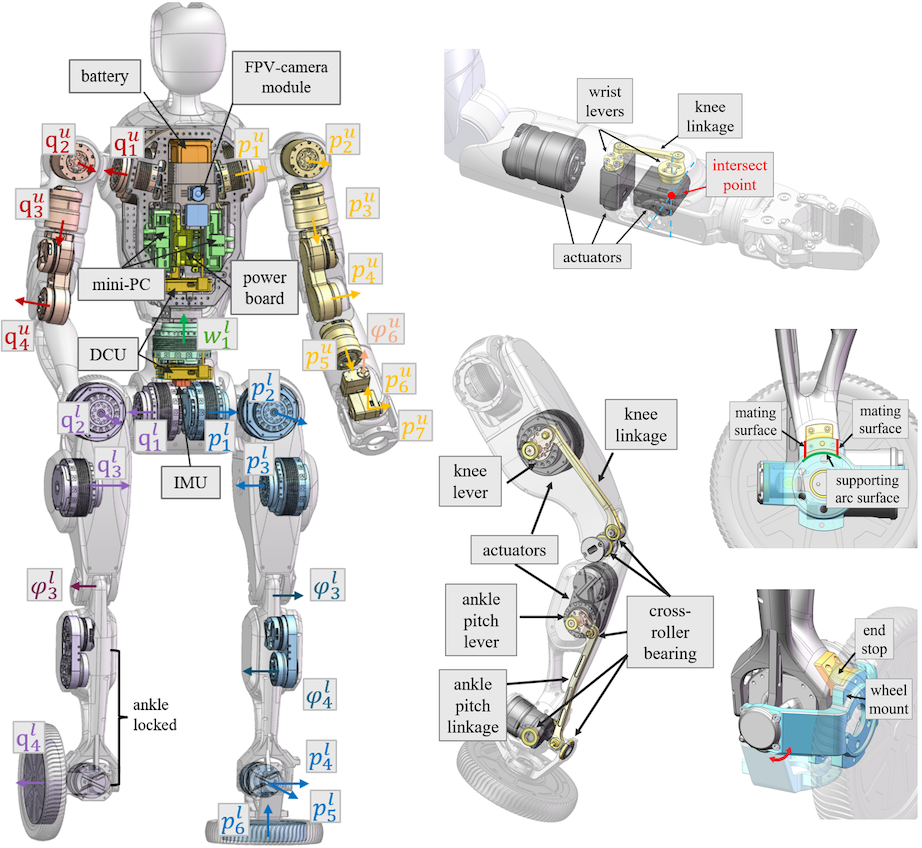}
    \caption{Illustration of X2-N's structural details (linkage and locking mechanism), and joint topology in different modes.}
    \label{fig:details}
    \vspace{-0.4cm}
\end{figure}

In foot-legged mode, the lower body consists of 13 DoFs, including a 1-DoF waist yaw joint and two 6-DoF legs. Each leg follows a conventional humanoid topology with a 3-DoF hip (pitch, roll, yaw), a 1-DoF knee (pitch), and a 2-DoF ankle (pitch, roll). Departing from standard designs, we introduce a topology-level reconfiguration by relocating the yaw joint. Specifically, the traditional flat foot is replaced by a wheel integrated with a direct-drive hub motor. Such motor is further reused as the yaw joint of the leg, effectively replacing the conventional yaw joint in the hip. By shifting leg yaw joint to the end-effector, X2-N integrates wheel-steering motors while preserving humanoid-compatible kinematics. The workspace and maneuverability of the proposed topology are analyzed along with a conventional humanoid leg in Section V, demonstrating a comparable operational capability. To further improve dynamic performance and motion agility, the knee and ankle pitch actuators are relocated closer to the CoM for reducing limb inertia \cite{Orin2013Centroidal, Sim2022TelloLeg}. We further design four-bar linkage mechanisms within the limbs for power and torque transmission of these joints.

In wheel-legged mode, as the yaw motion of X2-N can be achieved through wheel-velocity differentiation, the need for an explicit leg yaw joint is eliminated. Therefore, each leg is reconfigured to a 4-DoF topology including wheel steering. To omit redundant DoFs in leg of this mode, we design a structural locking mechanism and utilize the position control of ankle roll actuator to rigidly lock the wheel mount with the calf. Specifically, the mechanism has a retaining clip structure and adopt multiple contact interfaces to stabilize the wheel twist during locomotion, as shown in Fig. \ref{fig:details}. 

The mode transformation is achieved by reusing the ankle roll joint to actively drive the orientation of wheel motors from horizontal foot-contact to vertical wheel-contact form. Detailed transformation control process is described in the following controller section IV. Notably, during transformation, the locking mechanism can also adopt positioning and leading the ankle to the calf with a sliding slot structure. 

Through the design principle of joint reuse, topology reconfiguration and a locking mechanism, X2-N eliminates the need for additional dedicated actuators. Therefore, it achieves dual locomotion modes and transformation within a compact and efficient mechanical structure.

\subsection{Upper Limb Design}
A key design consideration for the arms is the trade-off between manipulation dexterity and dynamic agility. To balance both demands, we adopt a modular and interchangeable upper-limb design as illustrated in Fig. \ref{fig:details}, enabling the system to equip either configuration based on the requirements. 

In the basic configuration of X2-N, each upper limb provides 4 DoFs, including 2 DoFs in shoulder (pitch and roll), 1 DoF in upper arm (yaw), and 1 DoF in elbow (pitch). This configuration is designed for tasks such as body-balancing and payload delivery. The forearm is implemented as a rigid link for efficient motion and load bearing.

To enhance manipulation capability, the forearm can be replaced with a modular 3-DoF wrist unit (yaw, roll, pitch). The module can be installed at the elbow with minimal assembly effort. A linkage mechanism is employed to align the wrist joint axes to intersect in space, improving end-effector maneuverability. The arm can be equipped with a gripper or dexterous hand as end-effectors.

\subsection{Actuators}
\begin{table}[htbp]
    \centering
    \caption{Actuator specifications}
    \label{tab:actuator_spec}
    \small
    \setlength{\tabcolsep}{7pt}
    \renewcommand{\arraystretch}{1.3} 
    \begin{threeparttable}
    
    \begin{tabular}{lcccc}
        \toprule
        Actuator & R90 & R57 & R52 & R52-U \\
        \midrule
        Mass (g)          & 990 & 370 & 360 & 410 \\
        Gear ratio        & 16  & 40  & 36  & 72  \\
        Peak torque (Nm)  & 120 & 30  & 20  & 40  \\
        Peak speed (rpm)  & 105 & 110 & 130 & 65  \\
        \bottomrule
    \end{tabular}

\begin{tablenotes}[flushleft]
\footnotesize 
\item R90: shoulder pitch, hip pitch/roll, knee, waist;
R57: shoulder roll;
R52: arm yaw, ankle roll;
R52-U: elbow, ankle pitch.
\end{tablenotes}

\end{threeparttable}
    \vspace{-0.1cm}
\end{table}

To achieve high dynamic locomotion and manipulation performance, four types of customized actuators are developed, with specifications summarized in Table~\ref{tab:actuator_spec}. These actuators adopt quasi-direct-drive designs with planetary gear reductions, providing high torque density, low rotor inertia, and low rotational friction. Such characteristics enable a near-linear relationship between motor current and output torque, avoiding the need for extra torque sensors \cite{wensing}. High-torque actuators are assigned to load-intensive joints, including the shoulder, hip, and knee, while compact actuators are adopted for other joints to reduce the overall system weight.

\section{Control Framework}
\begin{figure}[t]
    \centering
    \vspace{0.15cm}
    \includegraphics[width=0.92\linewidth]{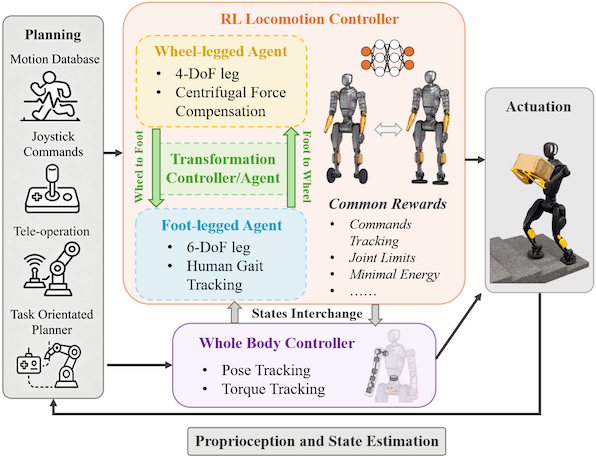}
    \caption{Illustration of X2-N's control architecture.}
    \label{fig:ctrl_arch}
    \vspace{-0.3cm}
\end{figure}

The control framework of X2-N integrates RL-based agents for locomotion and transformation, with a model-based whole-body controller for manipulation, as illustrated in Fig.~\ref{fig:ctrl_arch}

\subsection{Locomotion Controller}
The RL-based locomotion controller is realized through Proximal Policy Optimization (PPO)\cite{PPO} with an actor–critic architecture. We design three independent agents for locomotion and the controller can switch between them based on working scenarios, which are defined as follows: 
    \begin{itemize}
        \item \textbf{Wheel-legged mode}, which employs wheel propulsion and steering with the legs for posture adjustment.
        \item \textbf{Foot-legged mode}, which relies on legged stepping and turning with flat feet for locomotion.
        \item \textbf{Hybrid mode}, which combines the above behaviors to enable dynamic, skating-like motion, thereby fully exploiting the locomotion capabilities of X2-N.
    \end{itemize}

All policies are trained in the Isaac Gym simulator\cite{NVIDIA2021IsaacGym} and validated in the MuJoCo simulator\cite{mujoco2023} before real-world deployment. They share the same observation and action formulation. The observation vector consists of body pose, linear and angular velocities, joint positions, velocities, and torques. The actions are defined as desired joint position commands executed by the low-level joint controllers. 

Low-level actuation employs a torque–position–velocity hybrid controller following the MIT control policy \cite{wensing}. The control bandwidth is 50~Hz for learning policy inference, and 1~kHz for state estimation and actuator-level control.

\subsubsection{\textbf{Wheel-legged Agent with Centrifugal Compensation}}
Wheel-legged locomotion on X2-N presents significant challenges due to the high-DoF legs, heavy upper body, and elevated CoM, compared to conventional wheel-legged robots. These factors make it more difficult to maintain balance while achieving agile locomotion, especially during high-speed maneuvers and turning. To address this, we employ a two-stage RL training strategy. 

\begin{itemize}
    \item In \textbf{Stage A}, we first train the policy to learn fundamental wheel-based balance and velocity tracking on a flat terrain. We apply limited domain randomization and noise, and regularize joint motion to reference states. 

    \item In \textbf{Stage B}, the level of terrain complexity is progressively increased with uneven types, such as slopes, stairs, and low-friction surfaces to improve locomotion robustness. The reference joint states are removed to allow the policy exploiting coordinated motion of both wheels and high-DoF legs for adaptive locomotion.
\end{itemize}

\textbf{Centrifugal effects} substantially impacts locomotion stability of X2-N with elevated CoM during high-speed turning. Therefore, we introduce a body-leaning strategy that tilts the CoM toward the turning center, using gravity for partial compensation. During training, we design a centrifugal-force-aware reward term $R_{\text{cen}}$ based on a single rigid body model. The desired CoM tilt angle $\theta^{\text{des}}$ and the corresponding reward can be derived as:
\vspace{0.1cm}
\begin{align}
    \theta^{\text{des}} &= \arctan(\frac{v_x\omega_{\text{yaw}}}{g}), \\
    R_{\text{cen}} &= -(a_{y}^{\text{obs}} - \min{(0.3, \sin{\theta^{\text{des}}}))^2}, 
\end{align}
where $v_x$ and $\omega_{\text{yaw}}$ are the robot forward velocity and yaw angular velocity. $g$ is the gravity term, and $a_{y}^{\text{obs}}$ is the lateral-direction acceleration measured by the IMU. The constant 0.3 limits excessive body inclination for safety. This reward encourages the robot to generate appropriate body lean during turning for enhanced locomotion stability.

\subsubsection{\textbf{Foot-legged Agent with Human Gait Mimicry}}
The key distinction of foot-legged locomotion on X2-N compared to humanoid robots lies in its joint topology reconfiguration, where the hub motor is repurposed as the leg’s yaw joint. To achieve more bio-inspired motion and improve walking naturalness, we incorporate human gait data from the AMASS library \cite{mahmood2019amass} into the controller. To achieve these capabilities, we adopt a two-stage training process. 

\begin{itemize}
    \item In \textbf{Stage A}, the robot is trained on flat terrain to coordinate the yaw hub motor during stepping and turning, following desired commands. We apply a human-mimic reward $R_{\text{mimic}}$ to encourage the lower limb joints of X2-N to replicate human joint reference $\mathbf{q}_{\text{ref}}$. The reward  $R_{\text{mimic}}$ is defined as:
    \vspace{0.1cm}
    \begin{equation}
        \begin{aligned}
            R_{\text{mimic}} &= \exp(-2\|\mathbf{q} - \mathbf{q}_{\text{ref}}\|) \\
                        &- 0.2*\min(\|\mathbf{q} - \mathbf{q}_{\text{ref}}\|, 0.5).
        \end{aligned}
    \end{equation}
    \item In \textbf{Stage B}, the training environment is gradually diversified with stairs, slopes, and rough terrains, promoting robustness of high-dynamic locomotion. During this stage, the mimic reward is removed to prioritize locomotion performance. Nevertheless, the learned policy implicitly retains the human-like walking style.
\end{itemize}

\subsubsection{\textbf{Hybrid Wheel-legged Agent}}
Training hybrid wheel-legged controller is challenging due to the unsteady bipedal single-point contacts between the arc-shaped wheel surfaces and the ground, compared to flat-foot contacts. To address such challenges, we adopt a three-stage RL curriculum.

\begin{itemize}
    \item In \textbf{Stage A}, we train the robot with a frequent stepping cycle (T = 0.5 s) on flat terrains, faster than normal walking. By rapidly switching support points, the narrow-contact issue can be overcome. We further define a curve trajectory for the swing-leg end-effector. 
    \item In \textbf{Stage B}, the stepping duration is extended longer to 1.2s in each complete leg-switch cycle. We incorporated human walking data like foot-legged controller for bio-inspired gaits using the same reward $R_{\text{mimic}}$ in foot-legged agent.
    \item In \textbf{Stage C}, the training task becomes purely following body velocity commands without any reference trajectories. This allows the policy to develop adaptive stepping behaviors. We then gradually introduce rough terrains for robustness under high-speed locomotion. Notably, we penalize hub motor speed to zero at stepping contact instant to reduce jitter and vibration to the body. 
\end{itemize}

\subsection{Transformation Controller}
\begin{figure}[t]
    \centering
    \vspace{0.15cm}
    \includegraphics[width=0.95\linewidth]{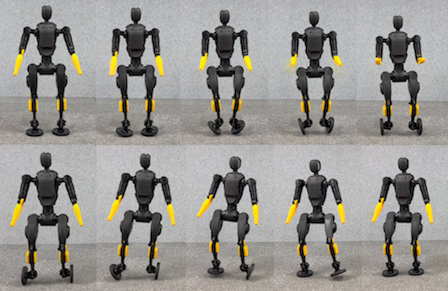}
    \caption{Illustration of X2-N's transformation processes. The top figures show the transition from foot-legged to wheel-legged; the bottom figures show the inverse mode transition.}
    \label{fig:transformation}
    \vspace{-0.4cm}
\end{figure}

The training setup of the transformation controller is similar as the locomotion controller. The core challenge lies in maintaining balance under inherently unstable changing ground contacts during smooth and robust mode transformation. To address this, we design two distinct transformation strategies for wheel-to-foot and foot-to-wheel transitions, illustrated in Fig. \ref{fig:transformation}.

\subsubsection{\textbf{From foot-legged mode to wheel-legged mode}} We apply a direct ankle twisting control to reorient the wheels from horizontal to vertical transitioning mode from foot to wheel. This approach follows the potential-energy-descent principle due to the CoM depression in wheel-legged mode. The mechanical components assistant with a rapid and precise ankle locking during the process. 

\subsubsection{\textbf{From wheel-legged mode to foot-legged mode}} We implement two distinct stepping motions while rotating the ankle and wheel during the air time of each step, transitioning wheel to foot. Since single-point contact is inherently unstable, the transformation duration is gradually refined to achieve smooth transformation and balance. 

During training of each process, offline joint trajectories for each transformation process are precomputed using forward kinematics. We assign each mode a fixed phase signal, and formulate the joint trajectories as a function of this signal and the transforming duration. The robot is then trained to maintain locomotion balance during transformation while following pre-defined joint trajectories. We specially define a balance reward $R_\text{balance}$ during the process:
\vspace{2pt}
\begin{align}
R_{\text{balance}} &= k_1 \exp\left(-\frac{\theta^2_{\text{roll}}}{\sigma_\theta^2}\right) + k_2 \exp\left(-\frac{\theta^2_{\text{pitch}}}{\sigma_\theta^2}\right)\\
& + k_3 \exp\left(-\frac{\omega^2_{\text{roll}}}{\sigma_\omega^2}\right) + k_4 \exp\left(-\frac{\omega^2_{\text{pitch}}}{\sigma_\omega^2}\right),
\end{align} to stabilize the body CoM and pose, where $k_1$ to $k_4$ are term weights, $\theta_\text{roll}, \theta_\text{pitch}, \omega_\text{roll}, \omega_\text{pitch}$ are the pose angles and velocities in roll and pitch orientations, and $\sigma_\theta, \sigma_\omega$ are the scale parameters to control decay rate.

\subsection{Whole-body Controller}
The whole-body controller is developed for task-oriented loco-manipulation, while the RL controller provides dynamic locomotion policies for balance stabilization and contact adaptation. The estimated robot states are shared with the RL agents as observations, and the whole-body controller generates commands in complement to RL-based actions.

For task-driven target poses, we define end-effector trajectories with pose, velocity, and force vectors in Cartesian space. The desired joint positions $\mathbf{q}$ and velocities $\mathbf{\dot{q}}$ are then computed via inverse kinematics and the end-effector Jacobian matrices $\mathbf{J_{\text{ee}}}$. Furthermore, the required joint torques $\mathbf{\tau}$ can be derived from external forces $\mathbf{F}_{\text{ext}}$ using the Newton–Euler dynamic equations:
\vspace{3pt}
\begin{equation}
\mathbf{M(q)}\mathbf{\ddot{q}} + \mathbf{C(q,\dot{q})} + \mathbf{g(q)} = \mathbf{\tau} + \mathbf{J_{\text{ee}}}^\top \mathbf{F}_{\text{ext}},
\end{equation} 
where $\mathbf{M(q)}, \mathbf{C(q,\dot{q})}, \mathbf{g(q)}$ represents the mass-inertia matrix, Coriolis, centrifugal and gravity terms. The derived joint positions, velocities, and torques are executed on the actuators via a low-level PD controller, following the MIT control policy \cite{wensing}.

\section{Simulation and Experiments}
The locomotion and operational capabilities of X2-N were validated through both simulation and experiments across diverse scenarios, demonstrating the effectiveness of the hybrid novel form and the robustness of the platform. 

\subsection{Workspace analysis}
\begin{figure}[t]
    \centering
    \vspace{0.15cm}
    \includegraphics[width=0.92\linewidth]{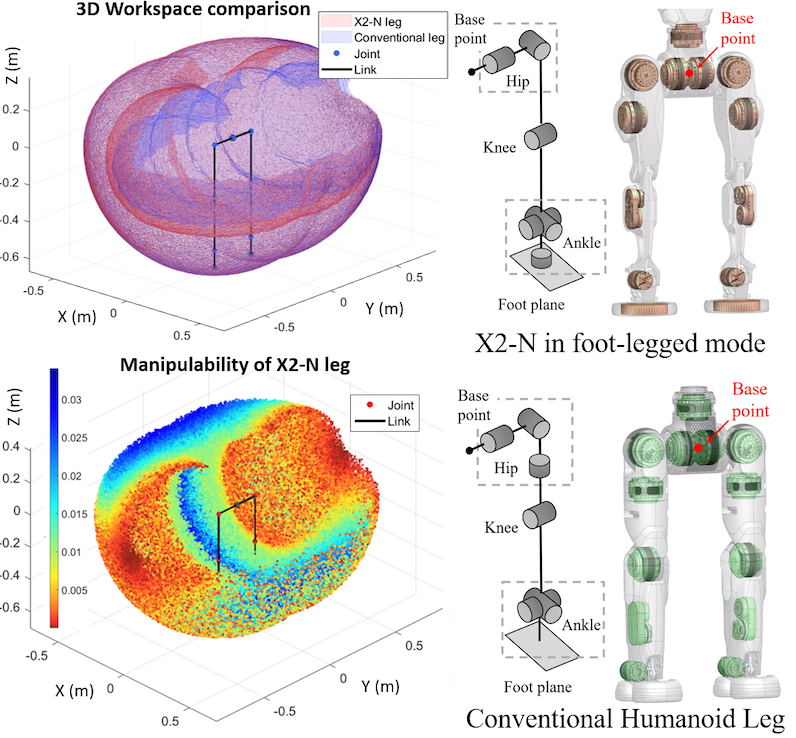}
    \caption{Leg-joint configuration comparison of workspace and manipulability between X2-N and conventional humanoid.}
    \label{fig:wrkspc}
    \vspace{-0.1cm}
\end{figure}

To evaluate the impact of the proposed joint reconfiguration in foot-legged mode, we compare the kinematic workspace and maneuverability of the X2-N's leg with that of a conventional humanoid leg. For a fair comparison, both configurations share identical thigh and calf lengths, and the illustration is shown in Fig. \ref{fig:wrkspc}. The corresponding Denavit–Hartenberg (DH) parameters and transformation matrices are derived to formulate the forward kinematics from the floating base (waist) to the end-effector (foot).

\begin{table}[htbp]
    \centering
    \caption{Workspace and maneuverability comparison}
    \label{tab:wrkspc}
    \small
    \setlength{\tabcolsep}{5pt}
    \renewcommand{\arraystretch}{1.4} 
    \begin{tabular}{lcccc}
        \toprule
        Configuration & X2-N leg & Normal leg & Ratio\\
        \midrule
        Volume (voxel) & 0.603 & 0.743 & 81.16\%\\
        Volume (convex hull) & 1.440 & 1.640 & 87.80\%\\
        Mean manipulability  & 0.0140 & 0.0154 & 90.91\%\\
        Max manipulability  & 0.0302 & 0.0359 & 84.12\% \\
        \bottomrule
    \end{tabular}
    \vspace{-0.1cm}
\end{table}

Considering joint limits, the reachable workspace of each configuration is sampled using a Monte Carlo method \cite{metropolis1949monte}, as illustrated in Fig.~\ref{fig:wrkspc}. The sampled points are used to construct workspace mesh surfaces, and the volume is evaluated using both convex hull coverage \cite{convex_hull} and voxel grid occupancy \cite{voxel} methods.

To further assess kinematic performance, we evaluate Yoshikawa’s manipulability index $w$ \cite{Yoshikawa1985Manipulability} for both configurations. For a consistent comparison, matched sampling points with the same spatial coordinates of both configurations are identified, and the mean manipulability $\bar{w}$ is computed as:
\begin{equation}
    \bar{w} =\frac{1}{N}\sum_{k=1}^{N} w_k =\frac{1}{N}\sum_{k=1}^{N}\sqrt{\det\!\left(\mathbf{J_{\text{ee}}}(\mathbf{q}_k)\mathbf{J_{\text{ee}}}^\top(\mathbf{q}_k)\right)},
\end{equation} 
where $N$ is the number of matched samples, and $\mathbf{J_{\text{ee}}}(\mathbf{q}_k)$ denotes the Jacobian matrix at $k$-th joint configuration $\mathbf{q}_k$.

The results show that the proposed topology achieves a workspace volume and mean manipulability comparable to those of a conventional humanoid leg (see Table \ref{tab:wrkspc}). Although a slight reduction in workspace is observed, the difference has minimal impact on foot placement in typical locomotion scenarios. These results indicate that the proposed joint reconfiguration preserves high locomotion capability while enabling additional functionality for transformation and wheel integration.

\subsection{Locomotion Performance and Energy Evaluation}

\begin{figure}[t]
     \centering
     \vspace{0.15cm}
     \includegraphics[width=1\linewidth]{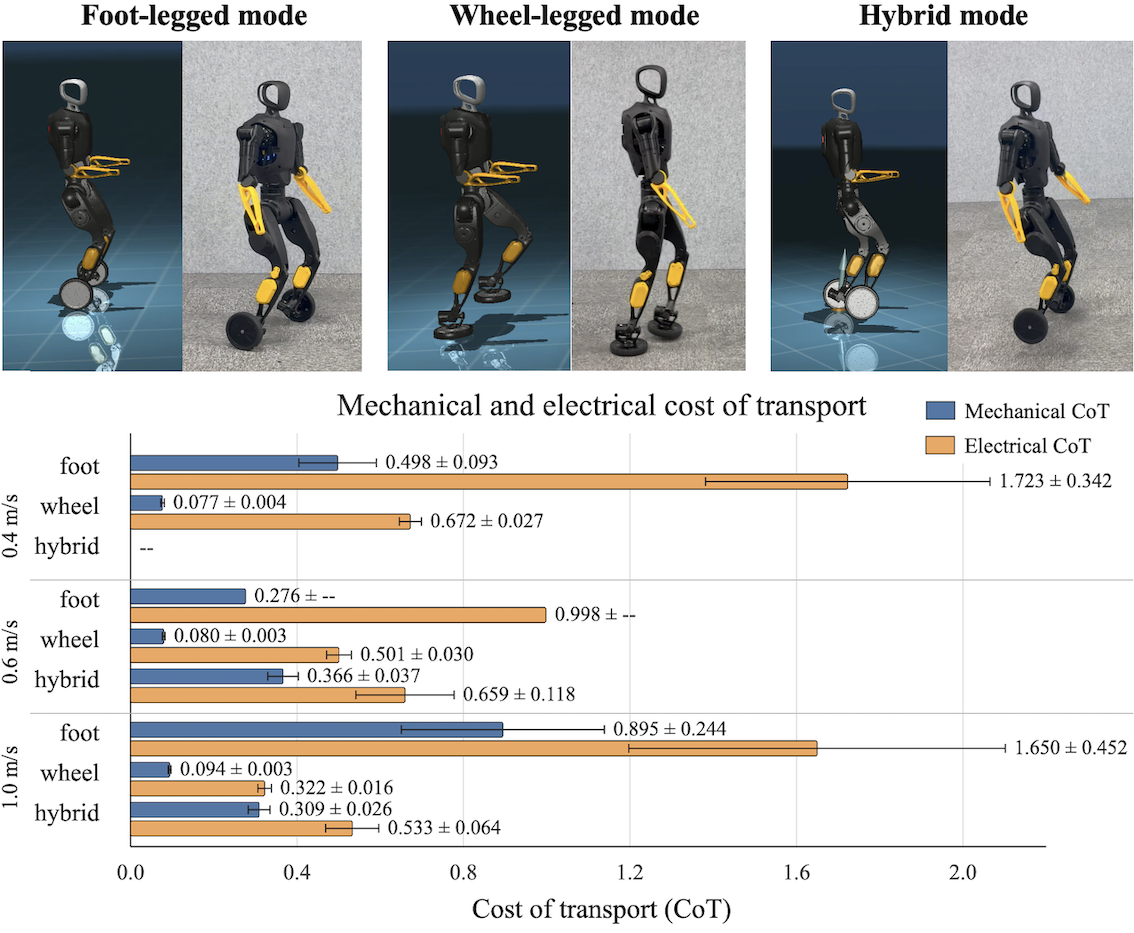}
     \caption{Three different locomotion modes of X2-N and average cost of transport.}
     \label{fig:3_cases}
     \vspace{-0.2cm}
\end{figure}

\begin{figure}[t]
    \centering
    \vspace{0.15cm}
    \includegraphics[width=0.97 \linewidth]{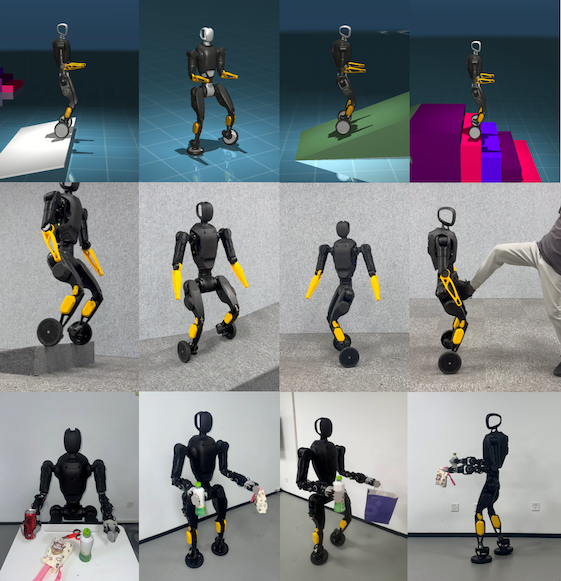}
    \caption{Experiments of X2-N on different scenarios.}
    \label{fig:sim_exp}
    \vspace{-0.5cm}
\end{figure}

Real-world locomotion performance was evaluated in wheel-legged, foot-legged, and hybrid modes, as illustrated in Figs. \ref{fig:3_cases} and \ref{fig:sim_exp}. In wheel-legged mode, we tested X2-N to traverse small ramps, perform in-place rotations on slopes, descend steps on stairs, and recover from external pushes. The legs act as an active suspension system, allowing traversal of highly uneven terrains. In foot-legged mode, X2-N behaves as a mid-size humanoid robot, performing anthropomorphic walking, stair and slope climbing, and push recovery. Moreover, we evaluated the hybrid wheel-legged locomotion mode, enabling agile and robust “skating" and “space-walking" behaviors.

Energy consumption was analyzed across different locomotion modes by calculating both mechanical and electrical power during locomotion cycles. The total energy consumption and cost of transport (CoT) formulas are as follows:

\begin{equation}
\begin{aligned}
E_{\mathrm{mech}}
&= \sum_{k=0}^{N-1}\sum_{i=1}^{n}
\left|\tau_i(k)\dot{q}_i(k)\right|\Delta t_k, \\[-2pt]
E_{\mathrm{elec}}
&= \sum_{k=0}^{N-1} V_k I_k \Delta t_k,
\; CoT = \frac{E}{mgd}.
\end{aligned}
\end{equation}
where $\tau_i(k)$ and $\dot{q}_i(k)$ denote the torque and angular velocity of joint $i$ at time step $k$, $V_k$, $I_k$ are the discrete voltage and current obtained from the battery management system, $d$ is the total travel distance.

\input{ener_result}

As illustrated in Table \ref{tab:three_mode_locomotion_power_energy}, legged locomotion experiences substantial impact power during each gait cycle, which can lead to structural fatigue and increased thermal load in the joints. In contrast, wheel-legged rolling can reduce joint peak load and energy consumption at equivalent forward speeds, demonstrating superior stability, efficiency, and robustness, and highlighting its practical advantages for diverse real-world constructed scenarios.

\subsection{Transformation and locking system Performance}
Transformation performance and repeatability were evaluated on constrained flat terrain, including both wheel-to-foot and foot-to-wheel transitions. The locking mechanism was further validated through mechanical analysis and hardware experiments with total 180 transformation cycles: 150 no-load cycles and 30 full-load cycles. We analyzed the transformation time $T_\text{trans}$, body position and orientation deviation $e_\text{CoM}, e_\text{ori}$, and the success rate.

From the results in Table~\ref{tab:transform_result}, each transformation was completed within approximately 2 seconds. During the transformation process, the robot maintained stable body posture with minor pose deviations. However, a small number of failure cases were observed, mainly attributed to mechanical component degradation after repeated operation.

\input{trans}

\subsection{Ablation Study of Rewards in RL-agents}
\begin{figure}[t]
    \centering
    \vspace{0.15cm}
    \includegraphics[width=0.95
    \linewidth]{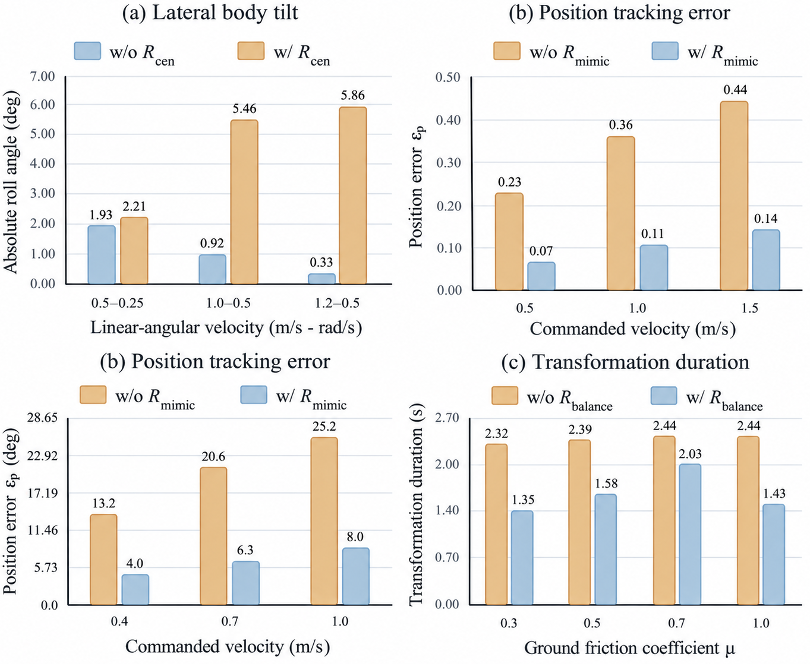}
    \caption{Ablation experiments results for key agent rewards.}
    \label{fig:abla}
    \vspace{-0.5cm}
\end{figure}

We further analyzed the contributions of the core reward components used in the RL agents. The ablation experiments demonstrate that the centrifugal-force compensation reward $R_{\text{cen}}$ improves turning agility and stability; the mimic reward $R_{\text{mimic}}$ reduces joint position errors with respect to the reference trajectories; and the balance reward $R_{\text{balance}}$ enables faster stabilization during the transformation process. Detailed experimental results are summarized in Fig. ~\ref{fig:abla} and the supplementary materials.

\subsection{Operation Performance}
We further assessed X2-N’s operational capabilities under both locomotion modes across diverse scenarios. Illustrations of different tests are shown in Fig. \ref{fig:sim_exp}. Leveraging its high-DoF upper limbs and grippers, X2-N can perform package carrying and object manipulation, while stationary and during locomotion, including operation on stairs and slopes. 

In stationary conditions, the foot-legged mode provides superior resistance to external forces and heavy payloads, maintaining steadier body and end-effector states. Conversely, during movement, the wheel-legged mode effectively mitigates oscillations induced by high-frequency ground impacts during legged walking, enabling smoother and more stable loco-manipulation performance.

\section{Conclusion and Future Work}
In this work, we presented X2-N, a novel transformable wheel-legged humanoid robot capable of dual-mode locomotion and manipulation. The proposed lower-limb joint topology reconfiguration and transformation mechanism enable seamless switching between wheel-legged and foot-legged modes without introducing redundant actuators. The workspace of X2-N's leg is preserved to be comparable to conventional humanoid legs. Moreover, the interchangeable arm design further enhances operational versatility. 

A RL-based whole-body control framework was developed and tailored to this morphology. It integrates multiple hybrid locomotion with manipulation behaviors and allows the robot to transform for the most suitable operation mode under different tasks. Extensive simulation and real-world experiments validate the advantages of X2-N with transformable design, demonstrating agility, efficiency and adaptability in locomotion as well as reliability in operation.

\bibliographystyle{IEEEtran}
\bibliography{references}

\end{document}


\begin{center}

{\LARGE\bfseries Supplementary Experimental Evaluation\par}

\vspace{0.6em}

{\large Supporting Material for the RA-L Manuscript:\par
\textit{“X2-N: A Transformable Wheel-Legged Humanoid Robot \\ with Dual-Mode Locomotion and Manipulation”}}

\vspace{1em}

{\large Related manuscript: arXiv:2604.21541
}

\end{center}

\vspace{0.5em}

\begin{figure}[H]
    \centering
    \vspace{0.15cm}
    \includegraphics[width=0.7\linewidth]{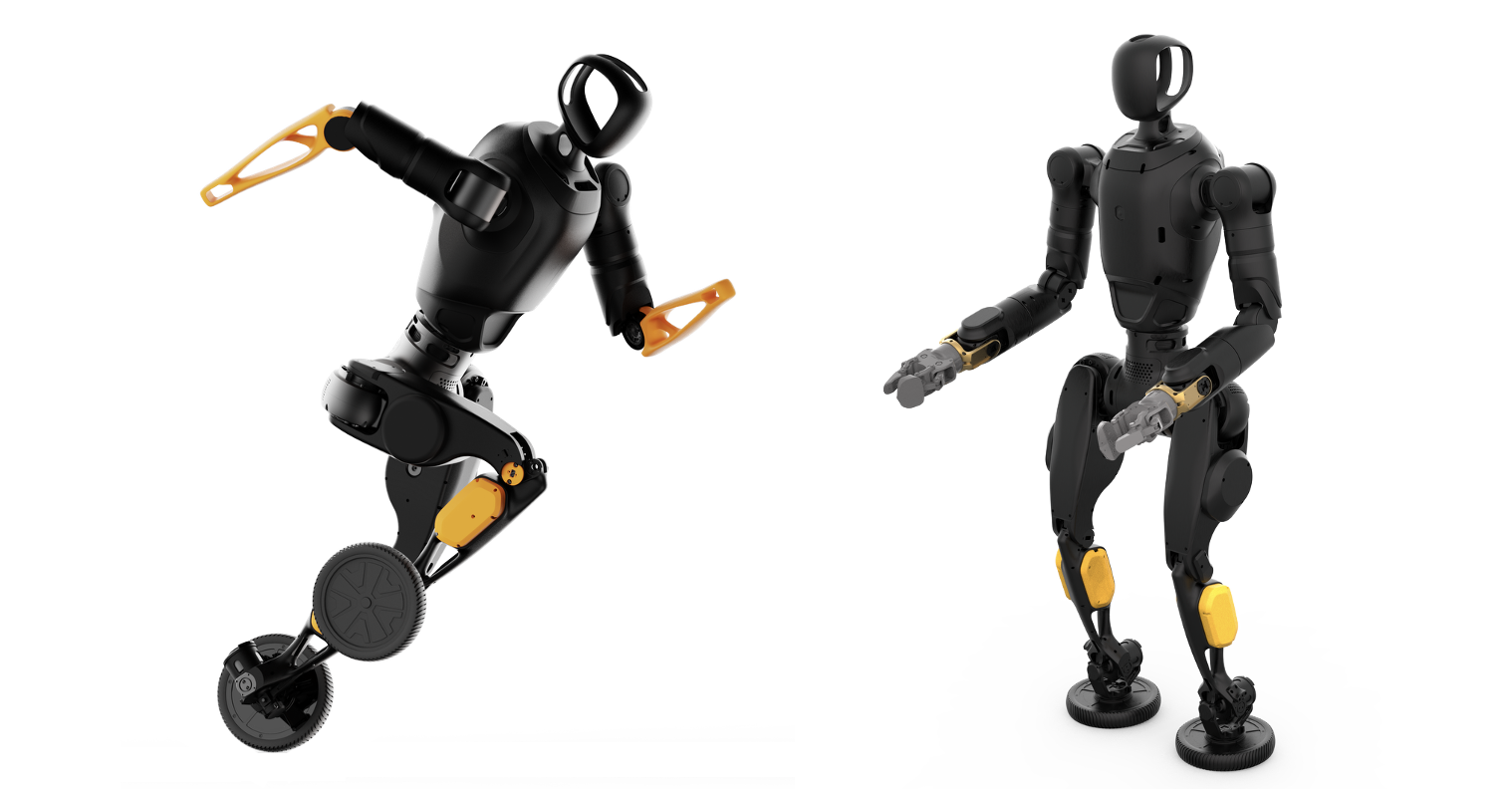}
    \caption{Overview of the X2-N platform with dual locomotion modes and different arm configurations.}
    \label{fig:overview}
    \vspace{-0.3cm}
\end{figure}

\section{Overview}

This supplementary document provides additional experimental evaluations 
to characterize the performance of the proposed X2-N transformable 
wheel-legged humanoid robot. Beyond the demonstrations presented in the 
main manuscript, these experiments provide more detailed quantitative 
metrics and assessments of its functional capabilities.

Furthermore, these evaluations address the major concerns raised by the 
reviewers regarding the core functionalities. The newly added experiments 
provide further quantitative evidence and supplementary analysis to 
strengthen the validation of the proposed system under the available 
experimental conditions. Some long-term engineering evaluations and 
additional test scenarios may be further explored in future work.

Five experiments are conducted, covering locomotion efficiency, 
transformation capability, locking system reliability, control performance, 
and workspace analysis, as summarized in Table~\ref{tab:exp_overview}. 
Detailed experimental setups, evaluation metrics, and results are presented 
in the following sections.
\medskip
\input{latex/tables/exp_overview}

\section{Fundamental Experimental Setup}

\subsection{X2-N Robot Specification}
The hardware specification of X2-N prototype is shown below as Table \ref{tab:robot_spec}. 
\medskip
\input{latex/tables/robot_spec}
\subsection{Environment Setup}
Both simulation and hardware environments were established for comprehensive performance evaluation.

\begin{figure}[H]
    \centering
    \vspace{0.15cm}
    \includegraphics[width=0.8\linewidth]{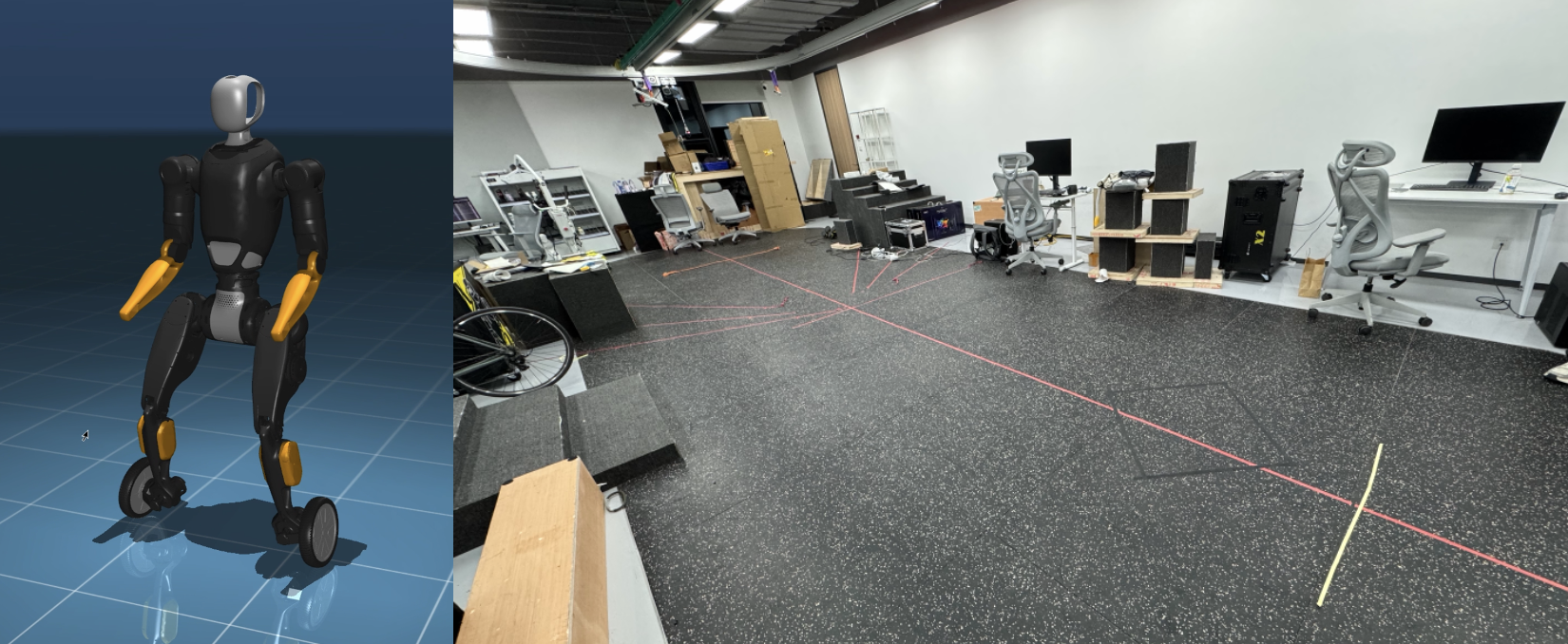}
    \caption{Overview of the simulation and hardware experimental environments.}
    \label{fig:overview}
    \vspace{-0.3cm}
\end{figure}

\medskip
\input{latex/tables/basic_envi_setup}

\subsection{Basic Data Acquisition}
During each experiment, the robot states and sensor measurements of 
X2-N were recorded for subsequent analysis, as summarized in 
Table~\ref{tab:basic_data}. For hardware experiments, the robot velocity 
and displacement were estimated from the measurements of the body-mounted 
IMU without external motion capture systems. The battery voltage and 
current were directly obtained from the battery management system (BMS).

\medskip
\input{latex/tables/basic_data}

\section{Experiment 1: Energy Efficiency Evaluation under Different Locomotion Modes}

To investigate the energy efficiency of different locomotion modes, 
additional simulation and hardware experiments were conducted on the 
proposed X2-N robot. The experiments quantitatively analyze the mechanical 
power, electrical power, total energy consumption, and Cost of Transport 
(CoT) of different locomotion configurations.

\subsection{Evaluation Metrics}
The evaluation metrics and corresponding formulas are summarized in 
Table~\ref{tab:ener_metric}. Since battery measurements are unavailable in 
the simulator, the electrical power cannot be estimated for simulation 
experiments.

\input{latex/tables/ener_metric}

\subsection{Experimental Design}
The experimental design for energy evaluation is described as follows. 
First, a static standing test is conducted to characterize the baseline 
energy consumption of the robot system, including the onboard computer, 
communication modules, electrical drive boards, and other auxiliary 
components. Then, X2-N is commanded to perform constant-speed straight-line 
locomotion under different locomotion modes to evaluate their energy 
efficiency. Detailed experimental settings are summarized in Table~\ref{tab:ener_setup}.

\begin{figure}[H]
    \centering
    \vspace{0.15cm}
    \includegraphics[width=0.8\linewidth]{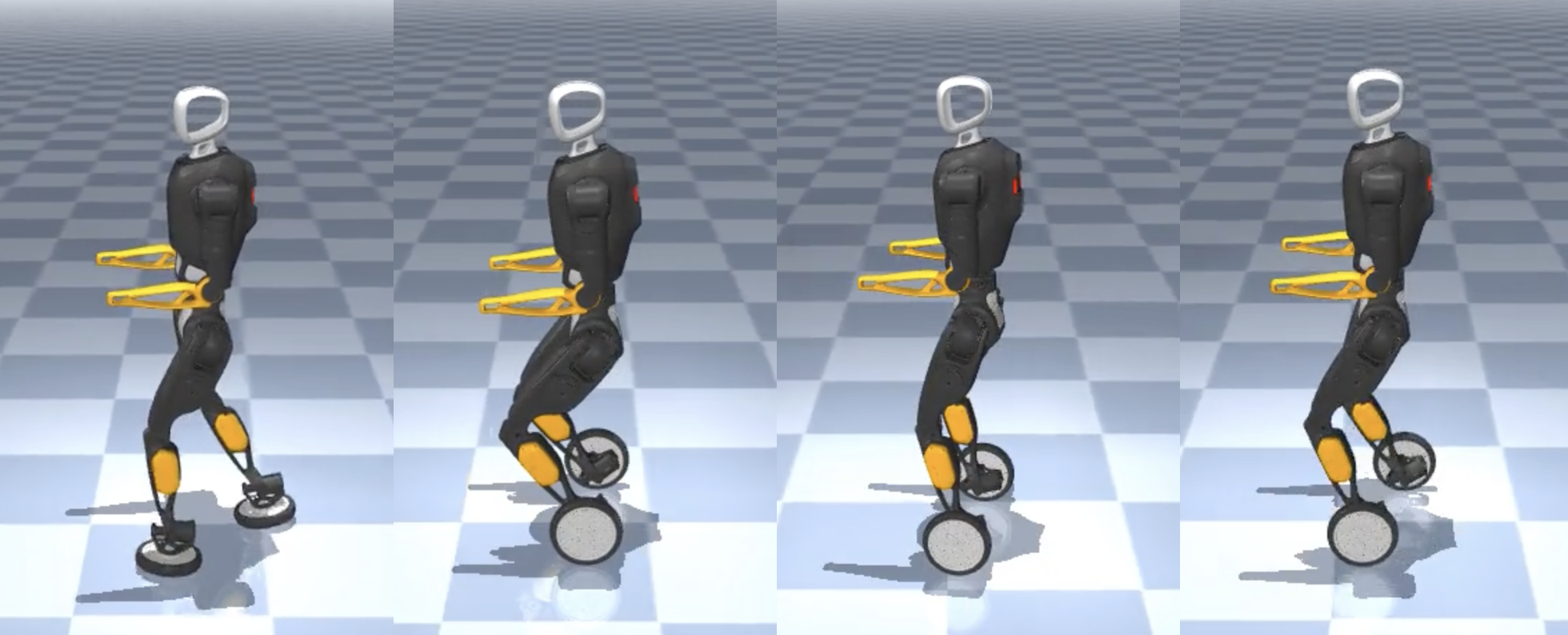}
    \caption{Four configurations in simulation setup.}
    \label{fig:overview}
    \vspace{-0.3cm}
\end{figure}

\begin{figure}[H]
    \centering
    \vspace{0.15cm}
    \includegraphics[width=0.8\linewidth]{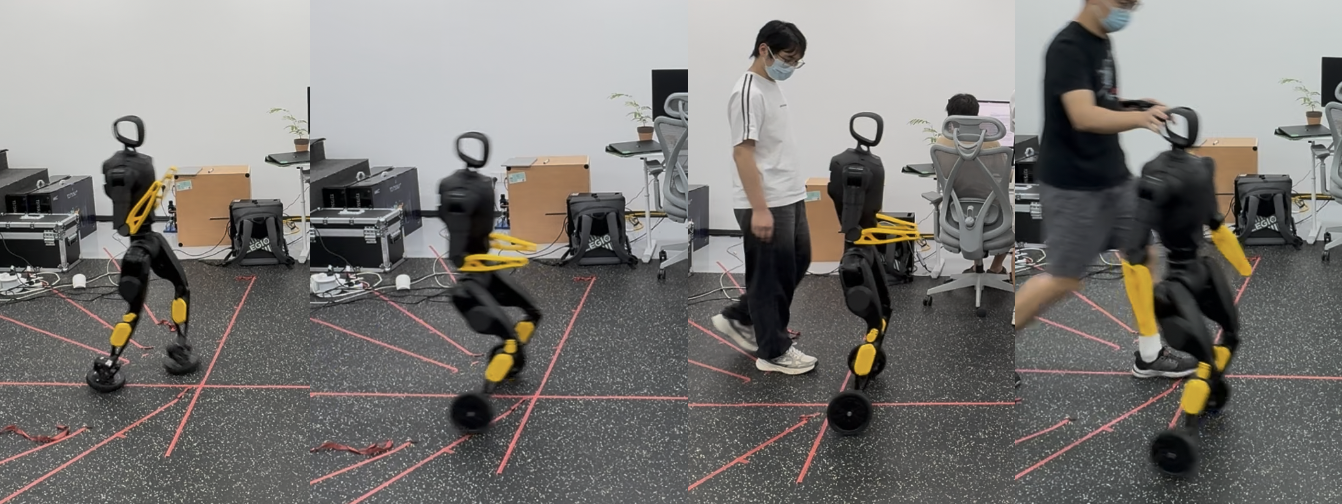}
    \caption{Four configurations in hardware setup.}
    \label{fig:overview}
    \vspace{-0.3cm}
\end{figure}

\input{latex/tables/ener_setup}

In addition, although the robot primarily relies on 
wheel locomotion in this mode, the bent knee posture introduces additional joint actuation and energy consumption. Therefore, the wheel-legged mode was further evaluated with a straightened-knee configuration (wheel-legged*). This additional case was 
conducted to analyze the potential energy consumption characteristics of the wheel-legged locomotion mode.

Furthermore, since the state estimation relies solely on onboard proprioceptive information (IMU), achieving accurate velocity tracking during experiments remains challenging. The actual velocity is influenced by multiple factors, including ground friction conditions, motion direction, sim-to-real gaps, joint temperature variations, and power supply stability. Therefore, the reported average velocities are determined based on practical hardware measured velocities.

\subsection{Results}
\subsubsection{Simulation}
\begin{figure}[H]
    \centering
    \vspace{0.15cm}
    \includegraphics[width=1\linewidth]{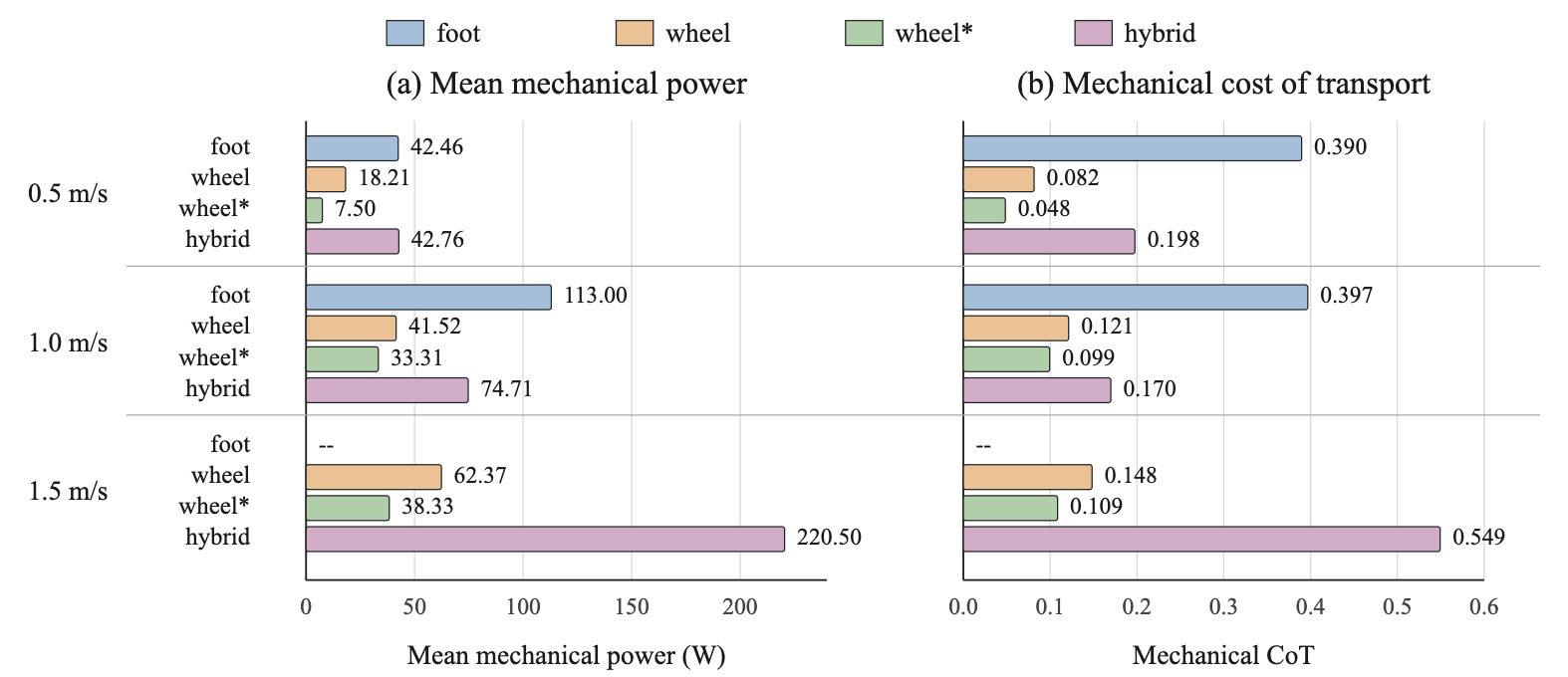}
    \caption{Mechanical power and CoT comparison in simulation experiments.}
    \label{fig:ene_simu}
    \vspace{-0.3cm}
\end{figure}

\begin{figure}[H]
    \centering
    \vspace{0.15cm}
    \includegraphics[width=0.8\linewidth]{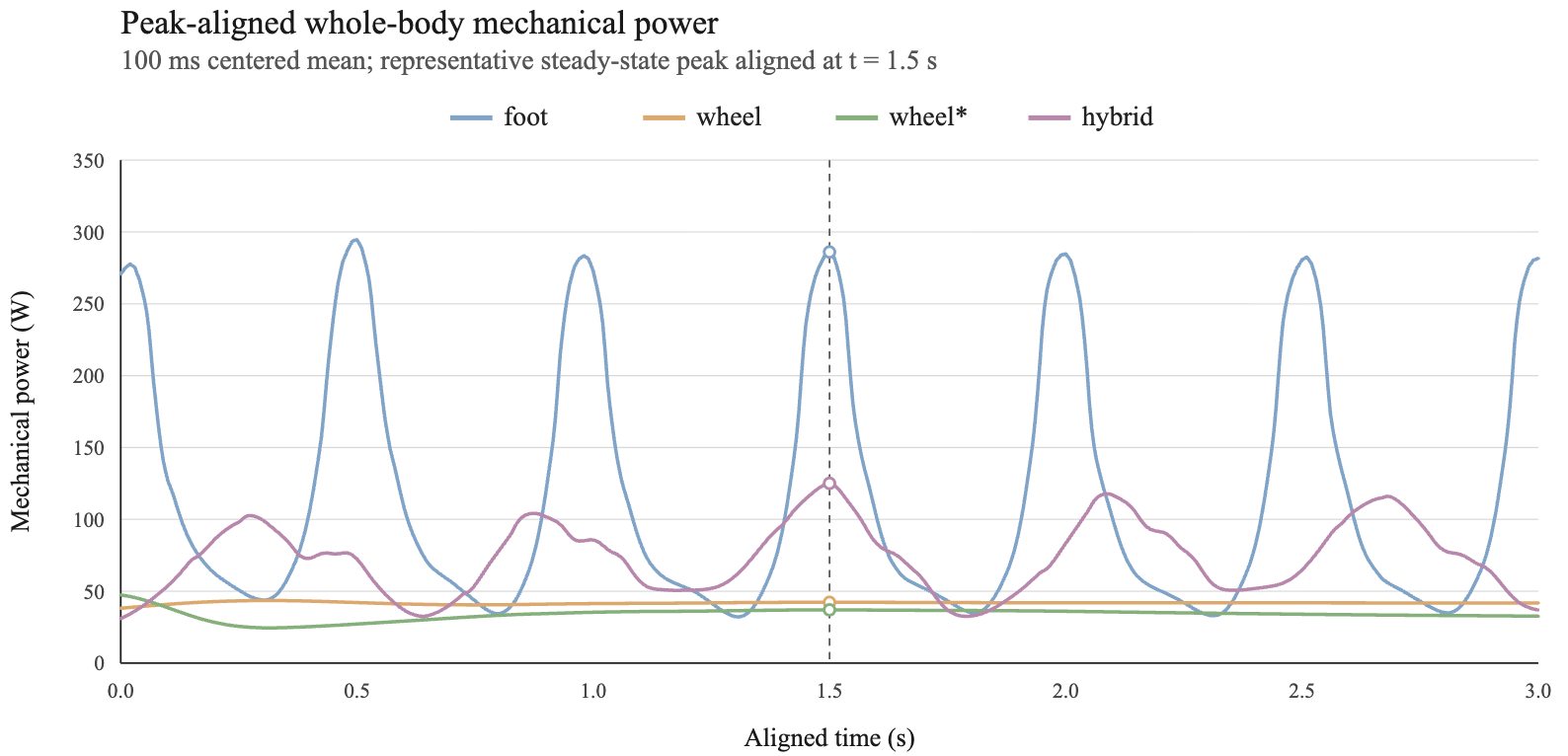}
    \caption{Real-time mechanical power in simulation experiments.}
    \label{fig:ene_simu3}
    \vspace{-0.3cm}
\end{figure}

Simulation results are shown in Fig. \ref{fig:ene_simu}, Fig. \ref{fig:ene_simu3} and Table \ref{tab:ener_simu_result2}. Due to joint torque limitations, the foot-legged mode cannot achieve a 
locomotion speed of 1.5 m/s, while the hybrid mode failed after 7 s of 
operation.

\input{latex/tables/ener_simu_result2}

\subsubsection{Hardware Experiment}
The hardware experiment statistics are provided below. It should be noted that the hybrid locomotion mode cannot maintain stable balance at low-speed conditions. Therefore, the 0.4 m/s experimental data are unavailable for this mode.

\input{latex/tables/ener_hw_result2}

\begin{figure}[H]
    \centering
    \vspace{0.15cm}
    \includegraphics[width=0.9\linewidth]{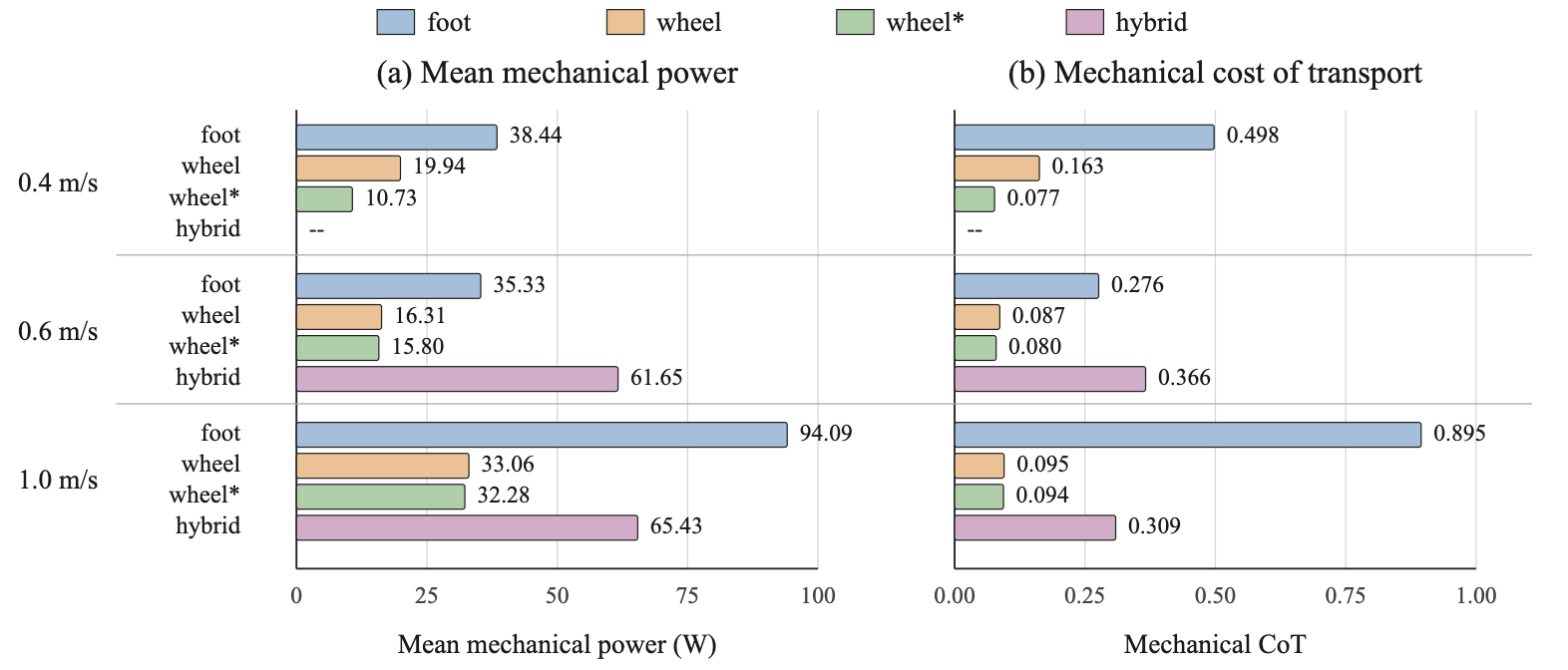}
    \caption{Mechanical power and CoT in hardware experiments.}
    \label{fig:overview}
    \vspace{-0.3cm}
\end{figure}

\begin{figure}[H]
    \centering
    \vspace{0.15cm}
    \includegraphics[width=0.9\linewidth]{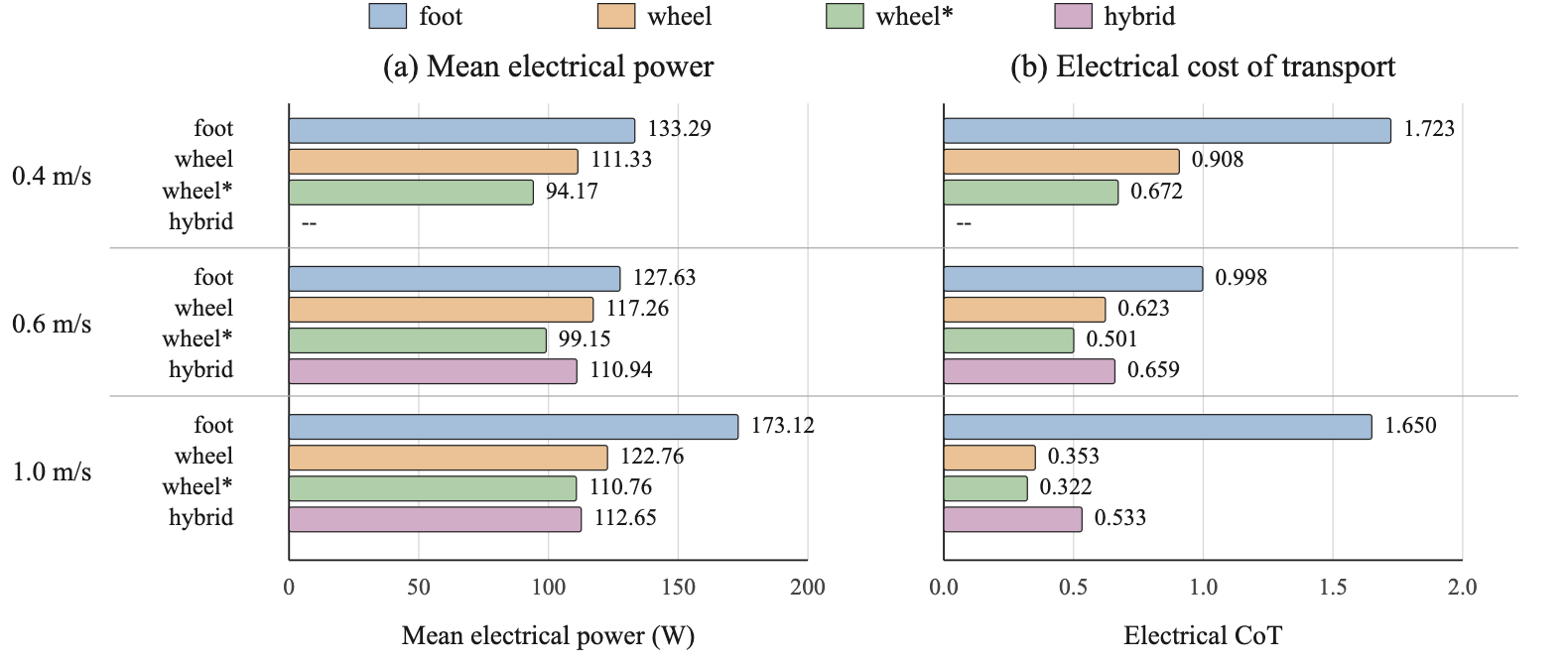}
    \caption{Electrical power and CoT in hardware experiments.}
    \label{fig:overview}
    \vspace{-0.3cm}
\end{figure}


\subsection{Discussion and Summary}
From the experimental results, wheel-legged locomotion achieves the lowest energy consumption under the same velocity conditions. This is mainly 
because the robot relies on continuous wheel rolling, which reduces the 
required joint actuation compared with foot-legged locomotion.

In contrast, foot-legged locomotion requires continuous joint actuation for 
gait generation, including leg swing motion, body stabilization, and 
periodic foot-ground interactions, resulting in higher mechanical and 
electrical power consumption. Hybrid locomotion exhibits intermediate 
energy consumption, as it retains the efficiency of wheel rolling while 
requiring additional leg joint actuation for posture regulation and motion 
coordination. 

Furthermore, the \textbf{straightened-knee} wheel-legged configuration further 
reduces energy consumption compared with the bent-knee configuration. This 
is because the straightened-knee posture improves structural load support 
and reduces the required joint torque, whereas the bent-knee configuration 
requires continuous actuator effort to maintain the posture.

And we found a \textbf{large sim2real gap} between the simulation and hardware experiments. The torque and velocity tracking errors are mainly because of sensing noise, ground friction conditions, motion direction, joint heat variations, and power supply stability.

Overall, the results validate the energy efficiency advantage of the 
transformable wheel-legged design. By switching to wheel-based locomotion 
when appropriate, X2-N significantly reduces actuator effort and achieves 
more efficient mobility.

\section{Experiment 2: Repeatability and Stability Evaluation of Bidirectional Transformation}

To further validate the engineering robustness of the proposed transformable mechanism, additional experiments were conducted to evaluate the repeatability, stability, and reliability of transformation between the wheel-legged and foot-legged configurations.


\subsection{Evaluation Metrics}

The following metrics are adopted throughout this section.
\input{latex/tables/transform_metric}

\subsection{Experimental Setup}
The transformation process consists of two operating modes: wheel-legged to 
foot-legged transformation and foot-legged to wheel-legged transformation. 
During each trial, the robot remains stationary while executing the 
transformation motion. The transformation completion time was determined based on the IMU measurements. When the body orientation signals reached a stable state after the transformation motion, the corresponding time was recorded as the transformation completion time.

To further evaluate the transformation robustness, hardware experiments were 
conducted under practical operating conditions. For the no-load condition, the 
robot was suspended and the ankle-driven transformation mechanism was 
activated to verify the basic transformation capability. For the full-body-load 
condition, a square region with predefined dimensions of 70 cm × 70 cm was 
marked on a flat floor to quantitatively evaluate transformation stability.

A successful transformation was defined as completing the mode transition 
without falling while keeping the robot body within the marked region. 
Otherwise, the trial was recorded as a failure. The corresponding experimental 
setup is illustrated in Fig. \ref{fig:trans_test}.

\input{latex/tables/transform_setup}

\begin{figure}[H]
    \centering
    \vspace{0.15cm}
    \includegraphics[width=0.7\linewidth]{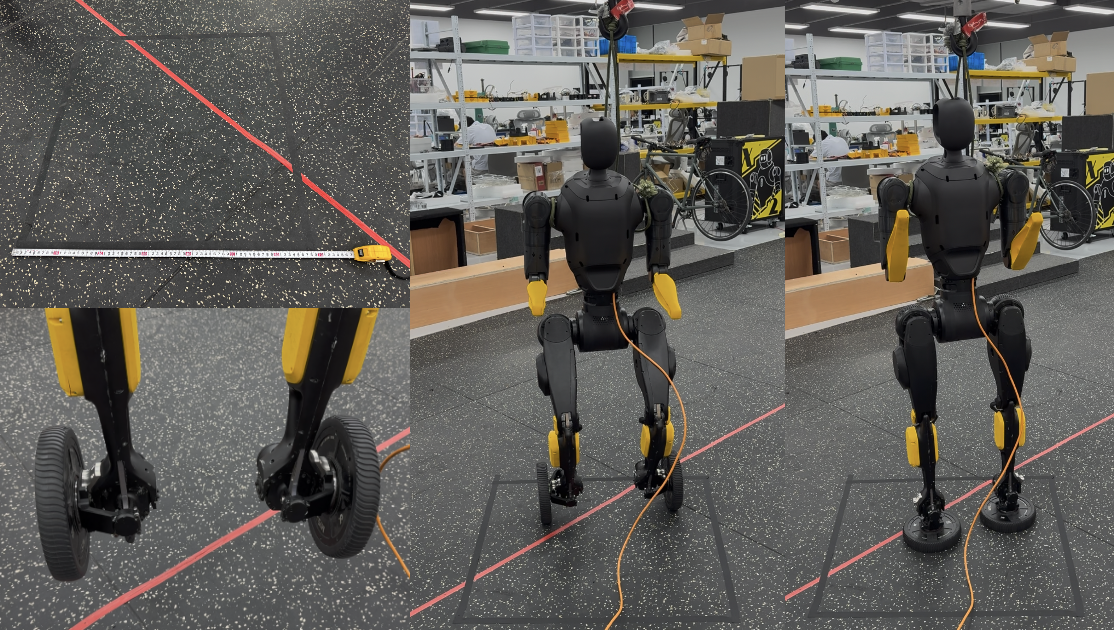}
    \caption{Transformation experiment setup.}
    \label{fig:trans_test}
    \vspace{-0.3cm}
\end{figure}


\subsection{Representative Results}
The recorded transformation data are summarized in Table~\ref{tab:trans_results}, Table \ref{tab:trans_results2} and Table \ref{tab:trans_success_rate} 
including the transformation time, CoM displacement, body orientation 
deviation, energy consumption, and transformation success rate.

\input{latex/tables/transform_result}
\input{latex/tables/transform_result2}
\input{latex/tables/trans_succ}


\subsection{Discussion}
Across all no-load transformation trials, the transformation controller and 
locking mechanism operated reliably without any jamming or structural damage. 
During the full-body-load transformation experiments, the robot maintained 
dynamic balance through continuous body posture regulation by the 
transformation controller. 

The proposed transformation and locking mechanisms 
achieved several repeated transformation cycles, demonstrating their 
practical reliability. The body displacement and orientation deviations remain within a relatively 
small range, especially during the foot-legged to wheel-legged 
transformation. 

However, several unexpected behaviors and failure cases were also observed during these experiments.

\paragraph{Yaw-axis Deviation} A noticeable yaw-axis deviation is observed during 
the wheel-legged to foot-legged transformation. This deviation occurs when 
the robot reaches the final stationary state and is mainly attributed to the 
current agent switching strategy in the control policy. As a result, the 
robot exhibits a consistent yaw rotation tendency toward the right side 
during this transformation direction. This issue will be further addressed 
through future controller optimization.

\paragraph{Failure Cases} 
Several failure cases were observed during the experiments, mainly caused by 
mechanical component degradation, joint overheating, operation errors, 
insufficient battery power, and unexpected embedded-system issues.

The \textbf{mechanical degradation} was mainly caused by loosened screws during repeated 
operation, which resulted in partial separation of the mechanical component 
from the bearing support. Consequently, the bearing lost its intended support 
function, and the reduced ankle structural stiffness introduced additional 
vibration, affecting the balance stability during transformation.

\begin{figure}[H]
    \centering
    \vspace{0.15cm}
    \includegraphics[width=0.6\linewidth]{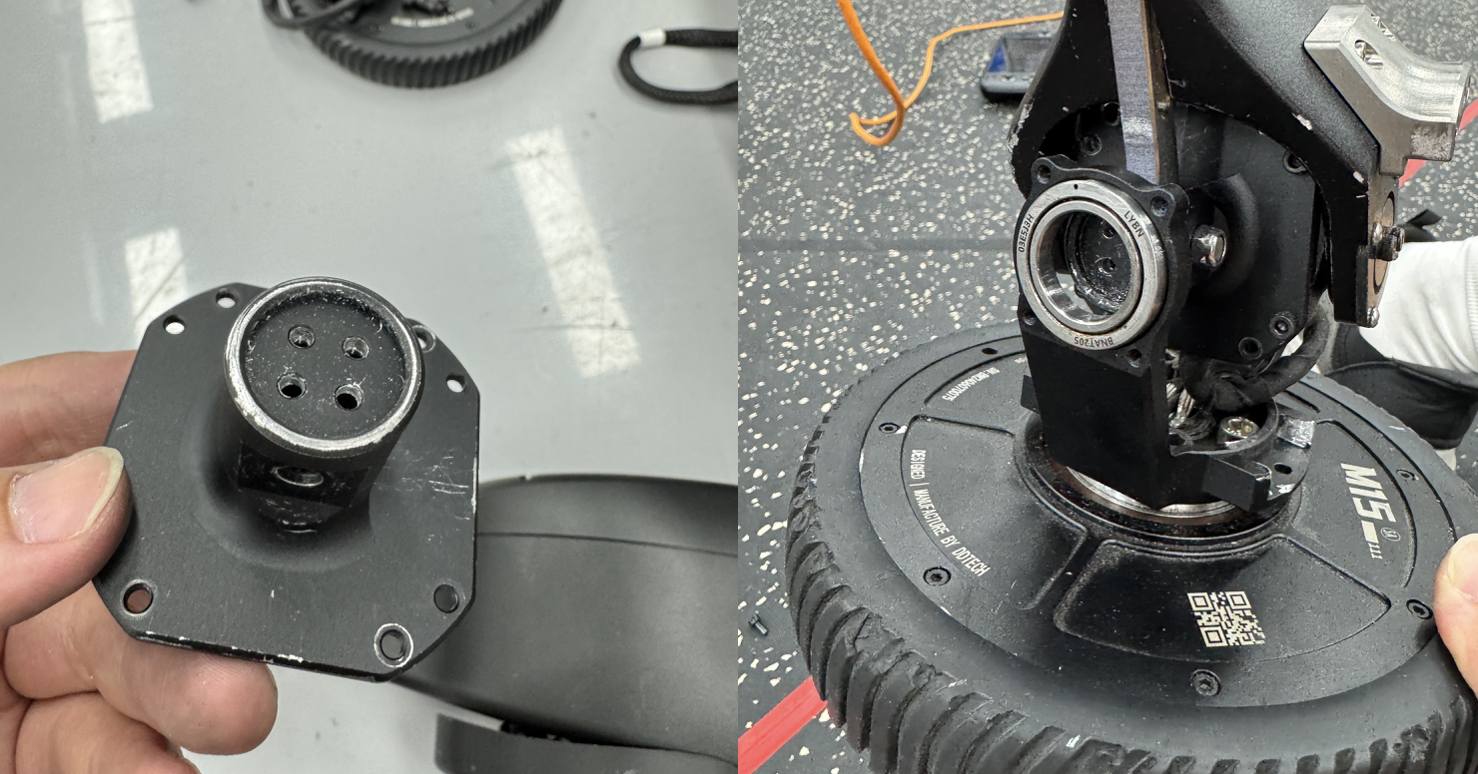}
    \caption{Mechanical degradation during transformation tests.}
    \label{fig:trans_test}
    \vspace{-0.3cm}
\end{figure}

Furthermore, continuous transformation requires sufficient \textbf{thermal management} 
and \textbf{battery power support} for the ankle actuators. Insufficient cooling and 
power stability reduced the achievable joint torque, leading to degraded 
posture control capability and occasional transformation failures.

\subsection{Summary}
The repeated transformation experiments validate the reliability and 
repeatability of the proposed transformation mechanism under practical 
conditions. The system maintains stable bidirectional transformation 
performance under different loading conditions, while the observed failure cases provide insights for further improvements in mechanical durability and controller robustness.

\section{Experiment 3: Engineering Validation of the Locking Mechanism}

The following experiments aim to evaluate the engineering performance of the proposed locking mechanism, including its structural strength, locking reliability, and operational stability under repeated transformations and external loading conditions.

\subsection{Detailed View of the Locking Mechanism}

\begin{figure}[H]
    \centering
    \vspace{0.15cm}
    \includegraphics[width=0.7\linewidth]{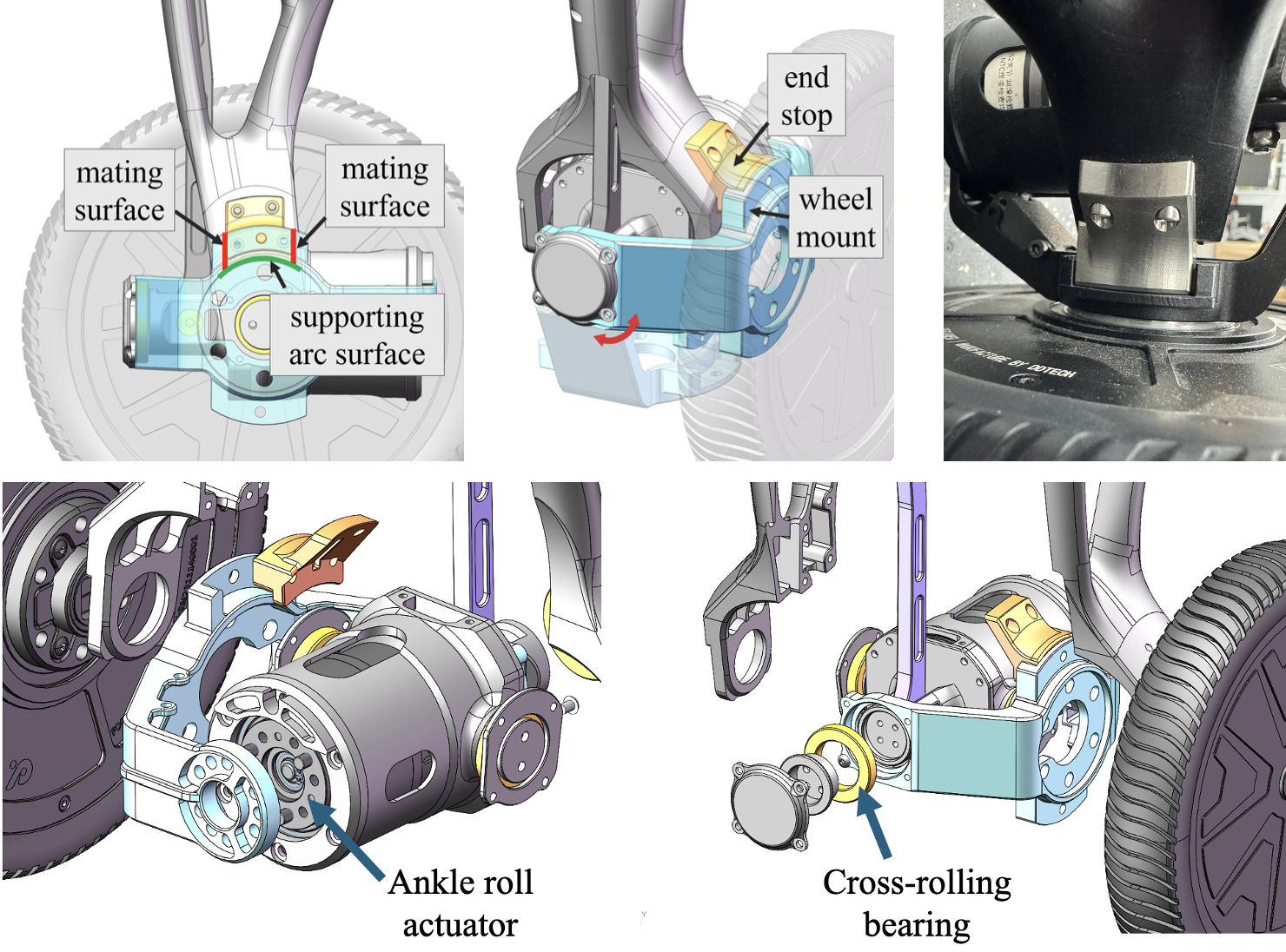}
    \caption{3D model and explosion view of the locking mechanism.}
    \label{fig:lock_model}
    \vspace{-0.3cm}
\end{figure}

\subsection{Evaluation Metrics}
The locking mechanism is evaluated using the following criteria.
\input{latex/tables/locking_metric}

\subsection{Experiment Setup}
The structural stiffness of the locking mechanism was evaluated using finite 
element analysis (FEA) in SolidWorks. The maximum loading condition of the 
ankle joint during transformation was first determined, and the corresponding 
load with an applied safety factor was then introduced into the simulation to 
assess the structural response and reliability of the locking mechanism.

\input{latex/tables/locking_setup}
\input{latex/tables/lock_load}

\subsection{Results and Discussion}
\subsubsection{Finite Element Analysis}
By assigning the material properties in the simulation model, the FEA results were obtained as shown below. The von Mises stress distribution was analyzed, where the red regions indicate the areas experiencing the highest stress.

The maximum simulated stress and deformation remain below the yield strength and allowable deformation limits of the aluminum 7075-T6 alloy material (approximately 500 MPa). Therefore, the locking components are expected to maintain sufficient structural reliability and robustness under the applied loading conditions.

\begin{figure}[H]
    \centering
    \vspace{0.15cm}
    \includegraphics[width=0.8\linewidth]{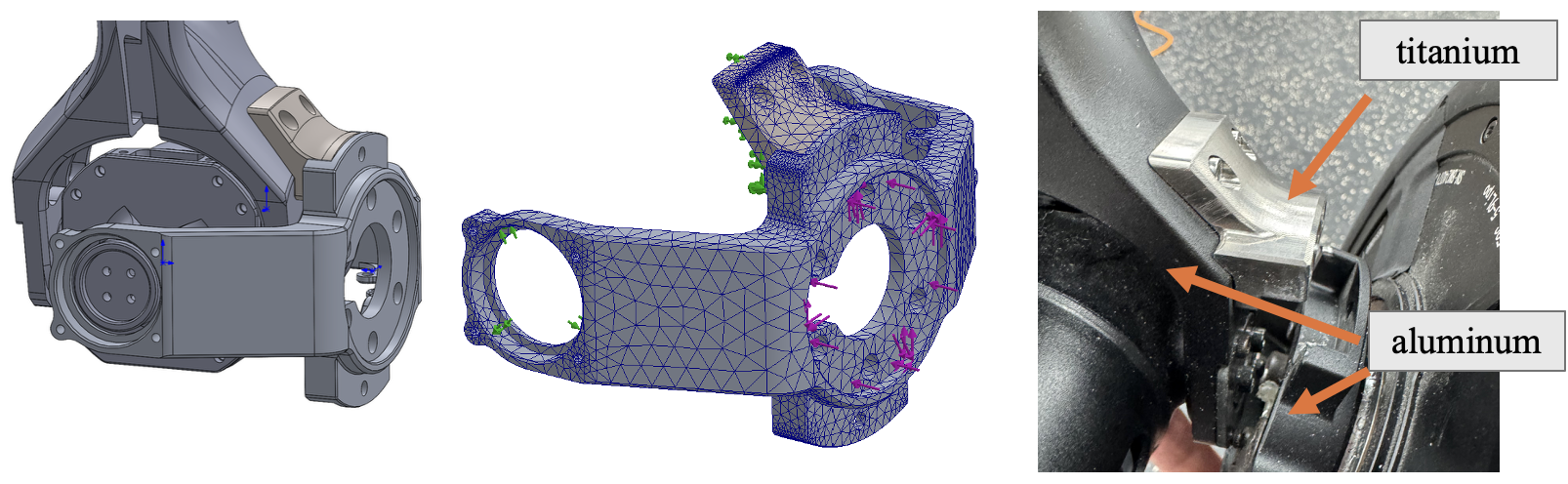}
    \caption{Modeling, mesh and real parts of the locking mechanism.}
    \label{fig:lock_mesh}
    \vspace{-0.3cm}
\end{figure}

\begin{figure}[H]
    \centering
    \vspace{0.15cm}
    \includegraphics[width=0.8\linewidth]{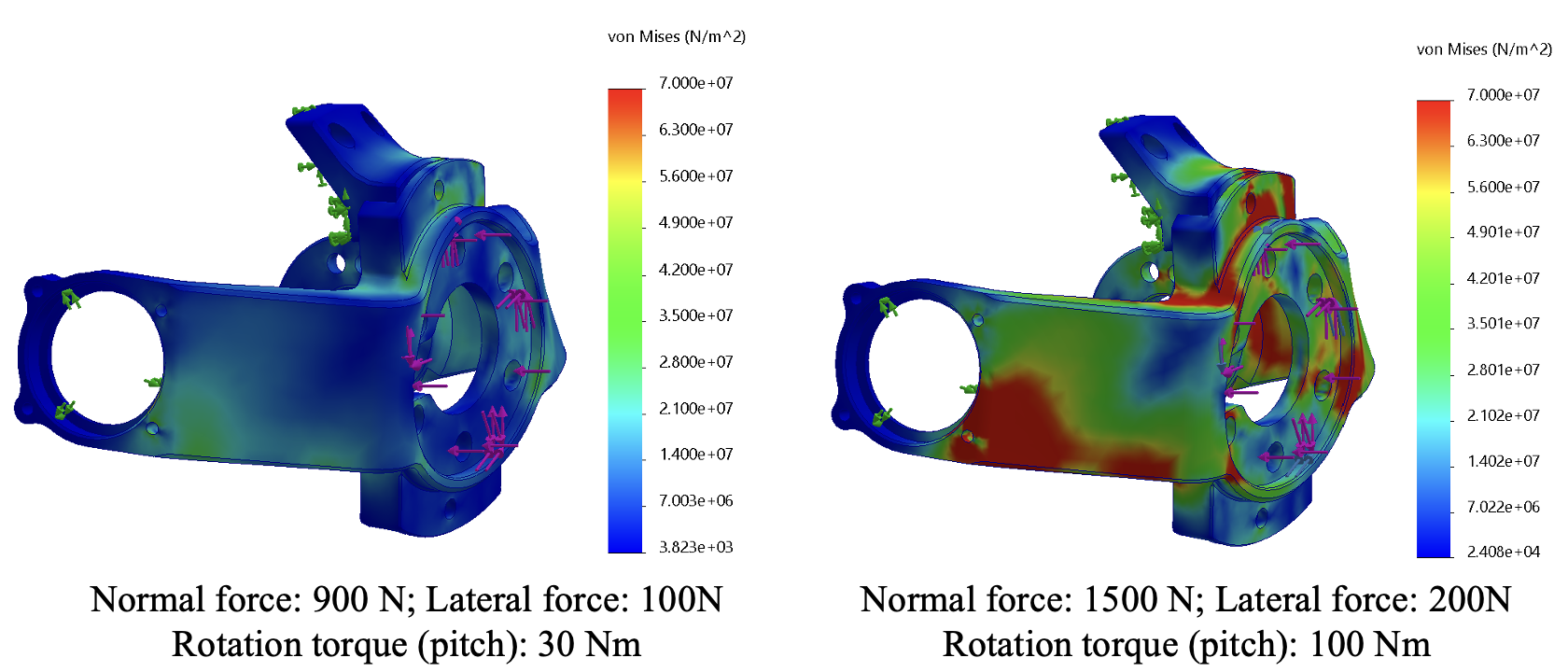}
    \caption{Stiffness analysis results of the locking mechanism.}
    \label{fig:lock_simu}
    \vspace{-0.3cm}
\end{figure}

\subsubsection{Backlash Analysis}
\begin{figure}[H]
    \centering
    \vspace{0.15cm}
    \includegraphics[width=0.7\linewidth]{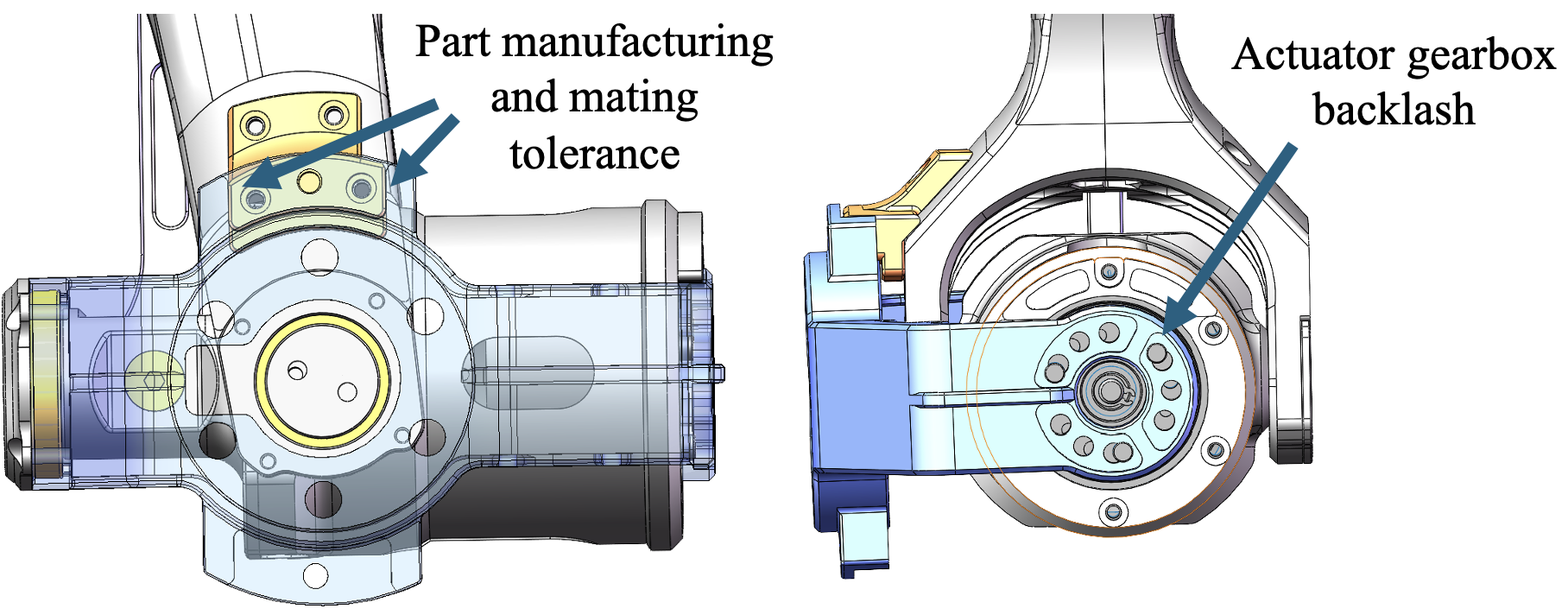}
    \caption{Pitch and roll view of the ankle of X2-N.}
    \label{fig:lock_backlash}
    \vspace{-0.3cm}
\end{figure}

We further analyzed the pitch and roll joint backlash based on the mechanical 
design and manufacturing tolerances.

For the \textbf{pitch rotation}, the joint is constrained by the mating surfaces of 
the locking components. Therefore, the dominant backlash source is the 
manufacturing tolerance. According to the 2D mechanical drawings, the 
machining tolerance of the metal CNC components is specified as 0--0.02 mm. 
With a rotation radius of approximately 30 mm, the maximum angular backlash caused by the clearance can be estimated as:

\[
\theta_{\mathrm{pitch,max}}
=
2*\arctan(\frac{0.02}{30})
\approx \textbf{0.076}^\circ \approx \textbf{4.56 arcmin}.
\]

For the \textbf{roll rotation}, the ankle joint is locked by the roll actuator. 
Therefore, the main backlash originates from the gearbox transmission. The 
R52 actuator adopts a planetary gearbox with an average backlash of 
approximately $\theta_{\mathrm{roll,max}} = \textbf{5 arcmin}$, corresponding to about \textbf{0.083}$^\circ$.

It should be noted that the backlash is not constant during the entire lifetime of the mechanism. With prolonged operation, mechanical wear and clearance accumulation can gradually increase the backlash. After extended usage, the backlash of the roll joint increased to approximately \textbf{50 arcmin}. The corresponding components will be replaced with newly manufactured parts during maintenance.

\subsubsection{Hardware Validation}
\paragraph{Experiments}
The hardware validation of the locking mechanism was performed through 
repeated transformation experiments. As mentioned in the previous section, 
we conducted 150 no-load transformation trials and 30 full-body-load 
transformation trials. The proposed locking mechanism achieved a high 
success rate throughout these experiments, demonstrating reliable locking 
performance under repeated transformation operations.

\input{latex/tables/trans_succ}
\paragraph{Wear Analysis}
After the initial assembly, the surface condition of the locking mechanism 
was smooth and intact, with an additional oxidation coating applied for 
surface protection. During subsequent controller validation and reliability 
testing, the robot underwent more than 80 full-body-load transformation 
cycles. During these repeated operations, different levels of mechanical wear 
were observed.

Compared with the titanium alloy components, the aluminum alloy components 
experienced more noticeable wear. The main wear region was located at the 
entrance area of the mating interface, which was mainly caused by repeated 
impact during the locking process and occasional failed transformations. The 
maximum observed wear depth was approximately $\Delta D = 0.5 mm$.

\begin{figure}[H]
    \centering
    \vspace{0.15cm}
    \includegraphics[width=0.6\linewidth]{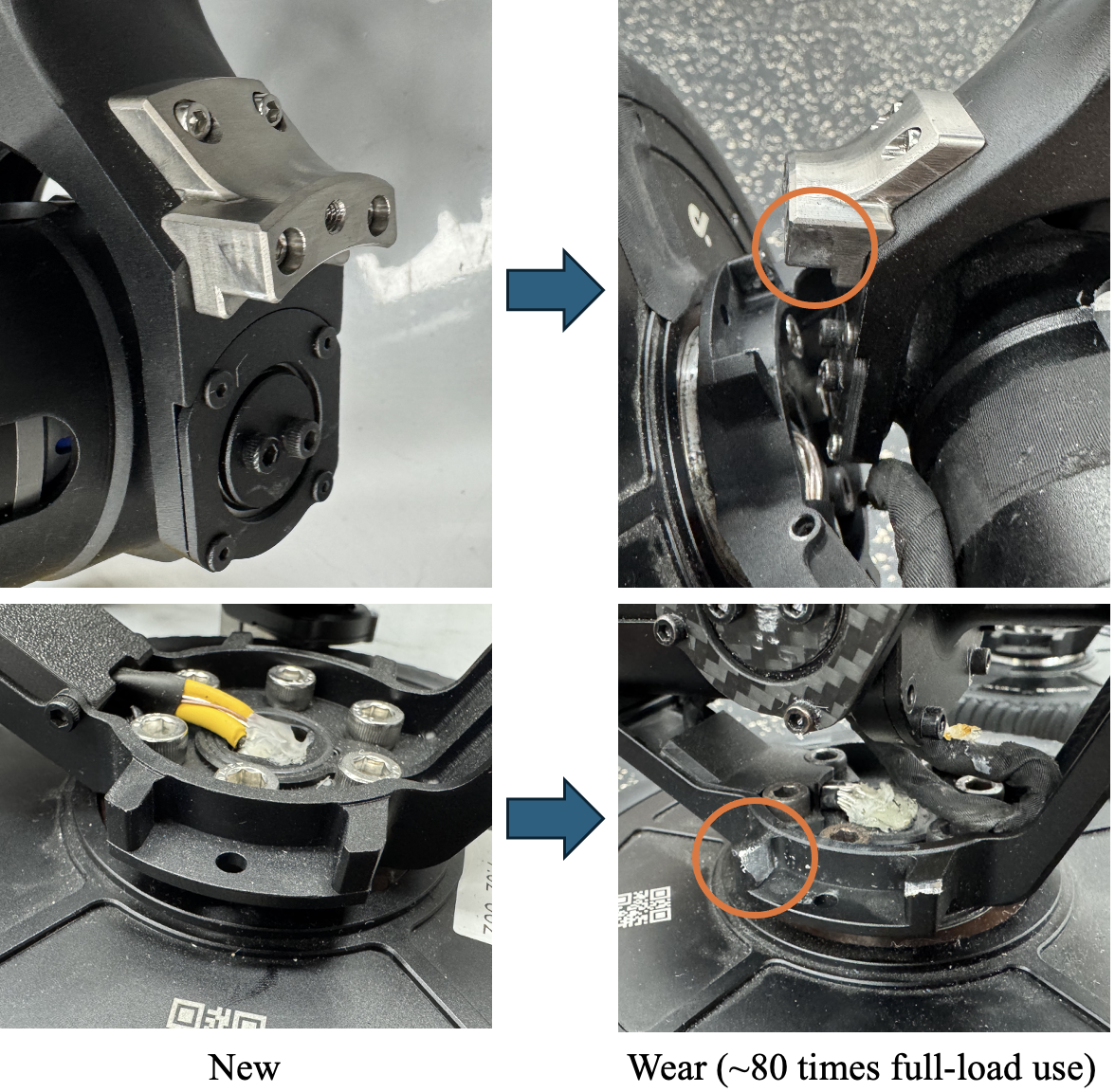}
    \caption{Wear cases of ankle parts.}
    \label{fig:wear}
    \vspace{-0.3cm}
\end{figure}

\subsection{Summary}
The locking mechanism was validated through repeated transformation trials 
under different loading conditions. The results demonstrate reliable locking 
performance during mode switching, while the observed wear mainly originated 
from repeated impacts at the mating interfaces. These findings verify the 
effectiveness of the proposed locking design and provide guidance for further 
durability improvements.

\section{Experiment 4: Ablation Studies of Reinforcement Learning Reward Components}
Based on the reward function used in the locomotion and transformation agents in the RL controller, we did three ablation research experiments to find the function of those rewards. 

The sub experiments are listed below, as shown in Table \ref{tab:ablation_exp_list}
\input{latex/tables/exp4_sub_overview}

\subsection{Centrifugal-force Compensation Reward Ablation}
\subsubsection{Reward Definition}
The centrifugal-force compensation reward encourages the robot to adopt an appropriate body-leaning strategy during turning. By exploiting the 
gravitational component generated by body inclination, the required 
centripetal force can be partially compensated, thereby improving turning 
stability. The corresponding mathematical formulation is given as follows:
\vspace{0.1cm}
\begin{align}
    \theta^{\text{des}} &= \arctan(\frac{v_x\omega_{\text{yaw}}}{g}), \\
    R_{\text{cen}} &= -(a_{y}^{\text{obs}} - \min{(0.3, \sin{\theta^{\text{des}}}))^2}, 
\end{align}
where $v_x$ and $\omega_{\text{yaw}}$ are the robot forward velocity and yaw angular velocity. $g$ is the gravity term, and $a_{y}^{\text{obs}}$ is the lateral-direction acceleration measured by the IMU. The constant 0.3 limits excessive body inclination for safety. This reward encourages the robot to generate appropriate body lean during turning for enhanced locomotion stability.

\subsubsection{Evaluation Metrics}
To capture the turning performance, we setup the follow evaluation metrics, shown in Table \ref{tab:anti_cen_metrics}.
\input{latex/tables/anti_cen_metric}

\subsubsection{Experimental Setup}
To evaluate the turning performance, X2-N was tested under 
circular motion with constant linear and angular velocities. An ablation 
study was conducted by removing the centrifugal-force compensation reward 
while keeping all other reward terms.
\input{latex/tables/anti_cen_setup}


\subsubsection{Results and Discussion}

\begin{figure}[H]
    \centering
    \vspace{0.15cm}
    \includegraphics[width=0.9\linewidth]{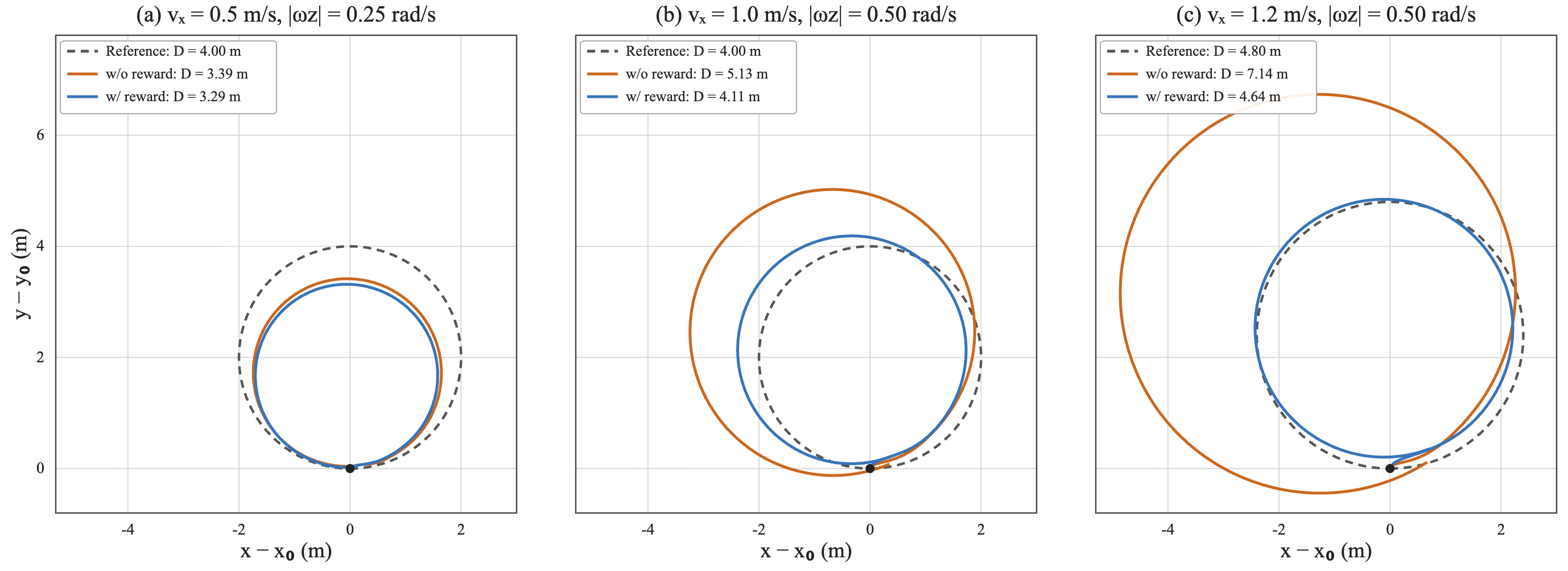}
    \caption{Circle trajectory of ablation studies.}
    \label{fig:overview}
    \vspace{-0.3cm}
\end{figure}

\input{latex/tables/anti_cen_result}

\begin{figure}[H]
    \centering
    \vspace{0.15cm}
    \includegraphics[width=0.9\linewidth]{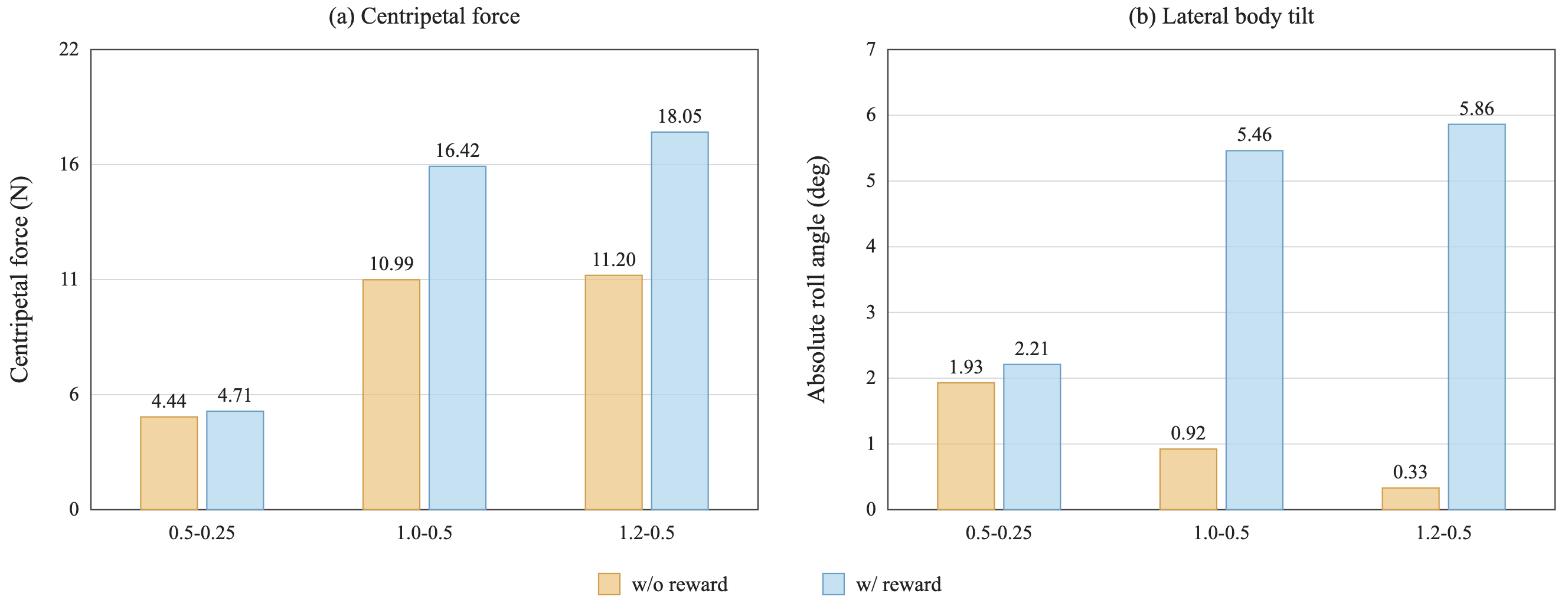}
    \caption{Comparison results of ablation studies.}
    \label{fig:overview}
    \vspace{-0.3cm}
\end{figure}

The ablation study demonstrates that the centrifugal-force compensation reward substantially improves circular-motion stability, particularly under demanding high-speed turning conditions. With the reward enabled, the robot maintains the commanded turning radius more accurately and exhibits improved velocity and centripetal-force tracking. The learned policy also develops a more pronounced body-leaning response, which helps preserve the intended turning behavior as the lateral demand increases.

At low speed, the benefit is less evident, and the reward does not consistently reduce roll-angle error. A modest increase in lateral slip is also observed in some conditions, suggesting a trade-off between geometric trajectory tracking and lateral motion regulation. Overall, the results indicate that the centrifugal compensation reward is most effective for maintaining stable and accurate turning at higher speeds, while further refinement is required to improve low-speed behavior and reduce residual lateral slip.

\subsubsection{Summary}
Through different speed experiments, it can be observed that removing the centrifugal-force compensation reward increases body roll and trajectory tracking error, especially during high-speed turning. The proposed reward effectively improves turning stability while reducing lateral slip, and help with wheel-legged locomotion stability.

\subsection{Foot-legged Walking Mimic Reward Ablation}
\subsubsection{Reward Definition}
We apply a human-mimic reward $R_{\text{mimic}}$ to encourage the lower limb joints of X2-N to replicate reference joint trajectory $\mathbf{q}_{\text{ref}}$. The reward  $R_{\text{mimic}}$ is defined as:
    \vspace{0.1cm}
    \begin{equation}
        \begin{aligned}
            R_{\text{mimic}} &= \exp(-2\|\mathbf{q} - \mathbf{q}_{\text{ref}}\|) \\
                        &- 0.2*\min(\|\mathbf{q} - \mathbf{q}_{\text{ref}}\|, 0.5).
        \end{aligned}
    \end{equation}

\subsubsection{Experimental Setup}
\input{latex/tables/mimic_setup}
The humanoid locomotion reward was removed while keeping all remaining reward terms unchanged. The controller was retrained using the same training environment. Simulation experiments were performed under constant-speed walking.

\subsubsection{Evaluation Metrics}
\input{latex/tables/mimic_metric}


\subsubsection{Results and Discussion}
\input{latex/tables/mimic_result} 
\begin{figure}[H]
    \centering
    \vspace{0.15cm}
    \includegraphics[width=1\linewidth]{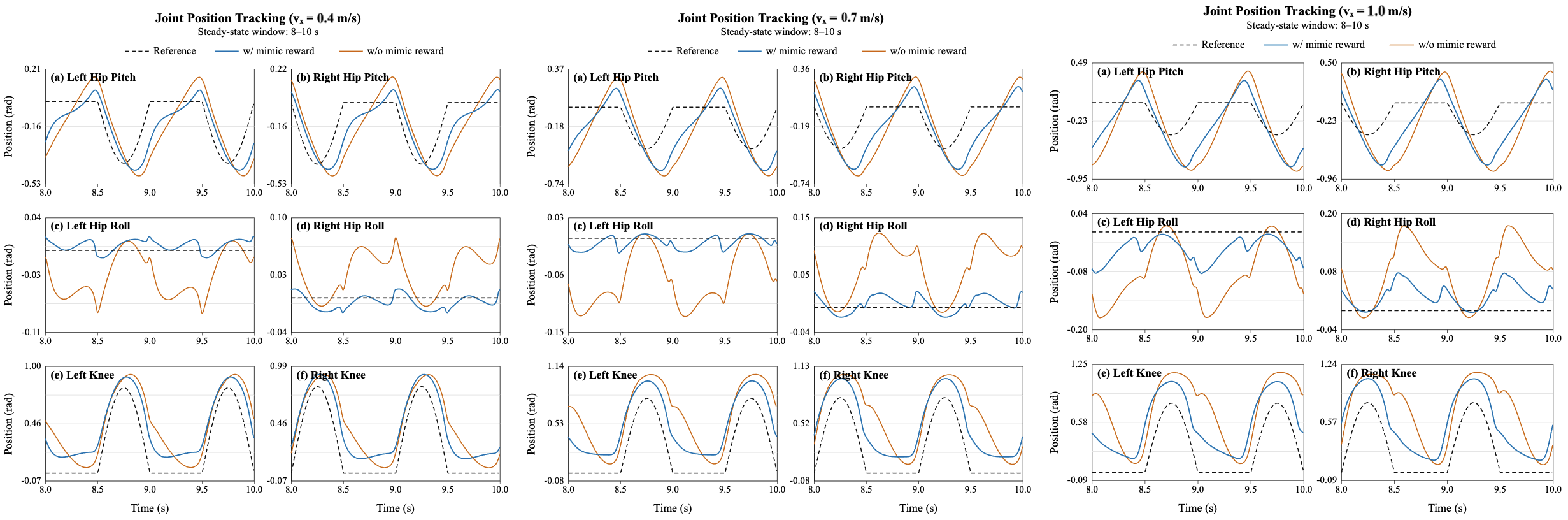}
    \caption{Joint position following errors of the ablation studies.}
    \label{fig:overview}
    \vspace{-0.3cm}
\end{figure}

\begin{figure}[H]
    \centering
    \vspace{0.15cm}
    \includegraphics[width=0.5\linewidth]{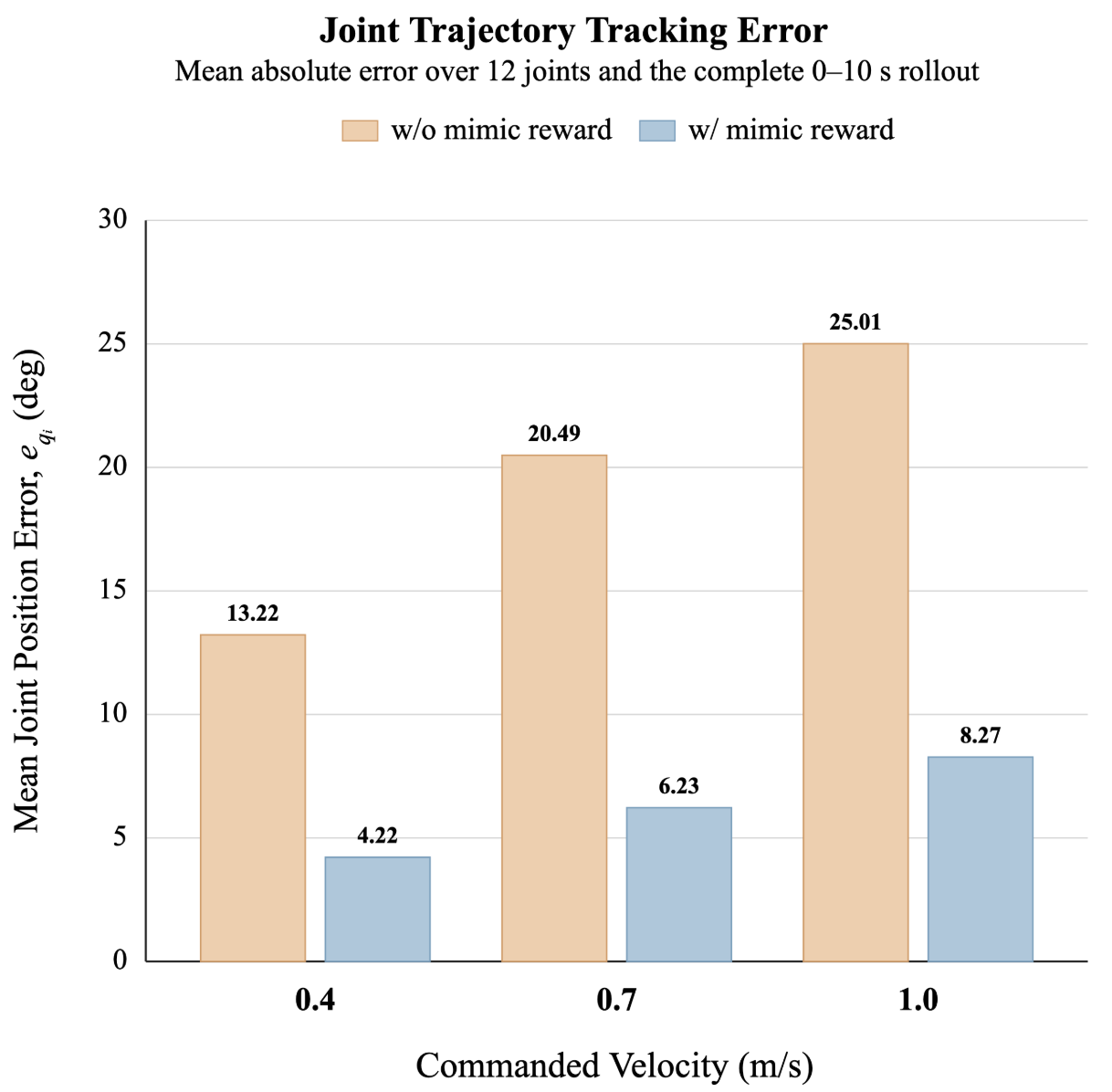}
    \caption{Comparison results of ablation studies.}
    \label{fig:overview}
    \vspace{-0.3cm}
\end{figure}

The mimic reward substantially improves joint-level motion tracking across 
all evaluated velocities, reducing the joint position error. The 
improvement is particularly pronounced for the ankle yaw and hip roll 
joints, while the improvement of knee joints is less significant at low 
walking speeds. However, improved joint trajectory imitation does not 
consistently translate into better whole-body stability. 

At higher walking 
speeds (0.7 and 1.0 m/s), the body orientation error increases, accompanied 
by larger yaw deviations. These results indicate that the mimic reward 
effectively enhances local joint-level imitation but introduces a trade-off 
between trajectory tracking accuracy and global body stability.

\subsection{Transformation Balance Reward Ablation}
\subsubsection{Reward Definition}
To maintain locomotion balance during transformation while following pre-defined joint trajectories, we specially define a balance reward $R_\text{balance}$ during the process:
\vspace{2pt}
\begin{align}
R_{\text{balance}} &= k_1 \exp\left(-\frac{\theta^2_{\text{roll}}}{\sigma_\theta^2}\right) + k_2 \exp\left(-\frac{\theta^2_{\text{pitch}}}{\sigma_\theta^2}\right)\\
& + k_3 \exp\left(-\frac{\omega^2_{\text{roll}}}{\sigma_\omega^2}\right) + k_4 \exp\left(-\frac{\omega^2_{\text{pitch}}}{\sigma_\omega^2}\right),
\end{align} to stabilize the body CoM and pose, where $k_1$ to $k_4$ are term weights, $\theta_\text{roll}, \theta_\text{pitch}, \omega_\text{roll}, \omega_\text{pitch}$ are the pose angles and velocities in roll and pitch orientations, and $\sigma_\theta, \sigma_\omega$ are the scale parameters to control decay rate.

\subsubsection{Experimental Setup}
\input{latex/tables/balance_setup}
The balance reward was removed while all remaining reward terms were retained. The controller was retrained using identical training settings.


\subsubsection{Evaluation Metrics}
\input{latex/tables/balance_metric}

\subsubsection{Results and Discussion}

\begin{figure}[H]
    \centering
    \vspace{0.15cm}
    \includegraphics[width=0.8\linewidth]{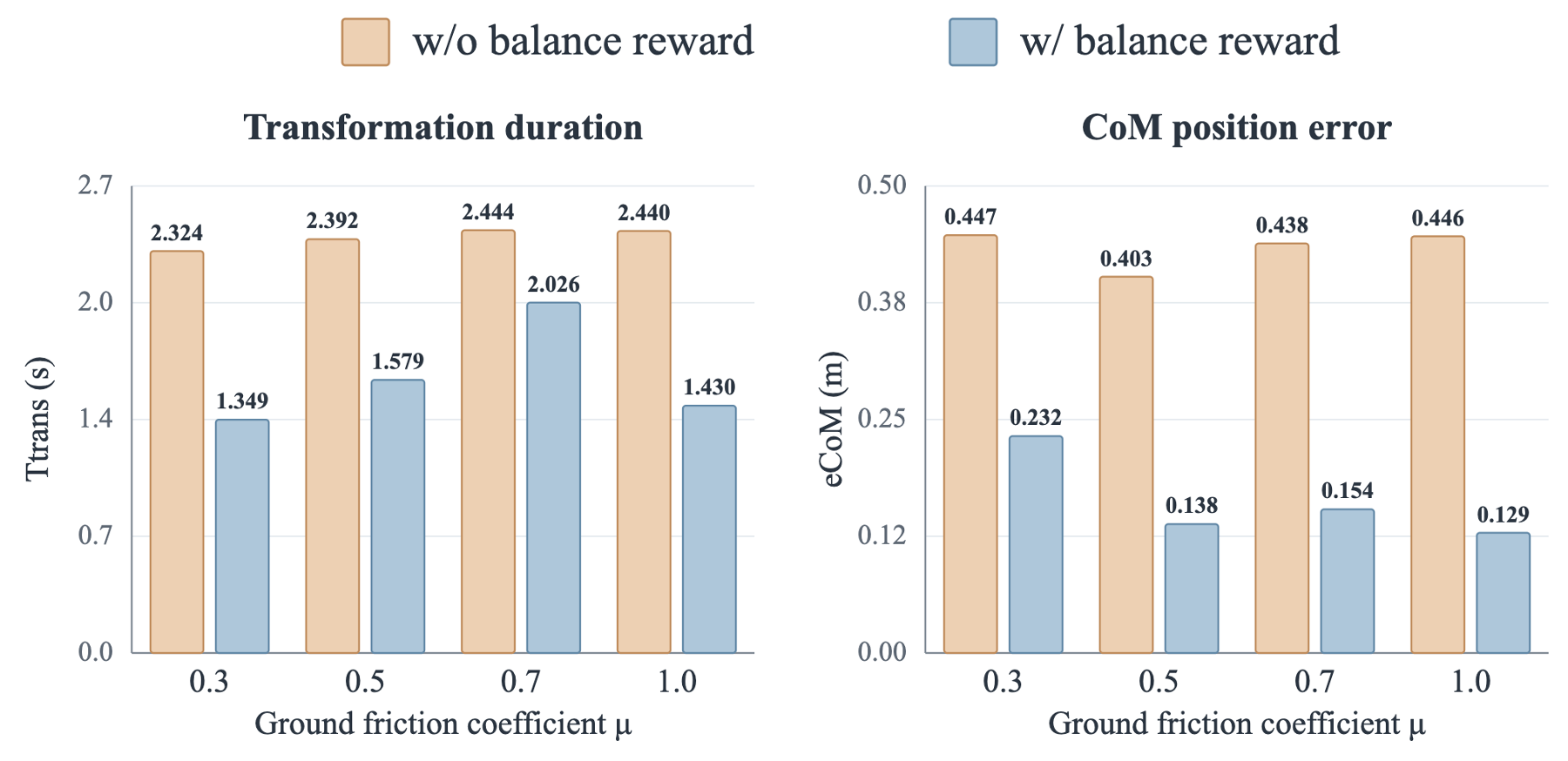}
    \caption{Comparison results of ablation studies.}
    \label{fig:overview}
    \vspace{-0.3cm}
\end{figure}

\input{latex/tables/balance_result}

Across all tested ground friction conditions, the balance reward consistently 
improved transformation efficiency and positional stability while maintaining 
reliable task completion. It enabled faster transformations and reduced CoM 
displacement. Meanwhile, a larger maximum body tilt angle was observed, 
indicating that the controller actively exploited body inclination to regulate 
the CoM and maintain balance during dynamic transformation.

These results demonstrate that the balance reward encourages a more dynamic 
balance strategy. Without the balance reward, the transformation controller 
tends to maintain a conservative upright posture but suffers from slower 
convergence and larger CoM deviation. With the proposed reward, the robot 
allows transient body attitude variations to achieve more effective balance 
regulation and robust transformation performance.

\subsubsection{Summary}

The ablation study verifies that the balance reward improves transformation 
stability by accelerating convergence and reducing body CoM deviation under 
different friction conditions. The results demonstrate its effectiveness in 
maintaining dynamic balance during morphology transformation.

\section{Experiment 5: Comparison Evaluation of 4-DoF and 7-DoF Interchangeable Arms}

\subsection{Overview}

To evaluate the practical benefits of the interchangeable arm design, we 
conducted a comparative analysis between the 4-DoF and 7-DoF arm 
configurations. The 4-DoF arm provides a lightweight and compact solution for basic manipulation tasks, while the 7-DoF arm introduces additional wrist degrees of freedom to improve end-effector orientation flexibility and 
dexterity.

\begin{figure}[H]
    \centering
    \vspace{0.15cm}
    \includegraphics[width=0.65\linewidth]{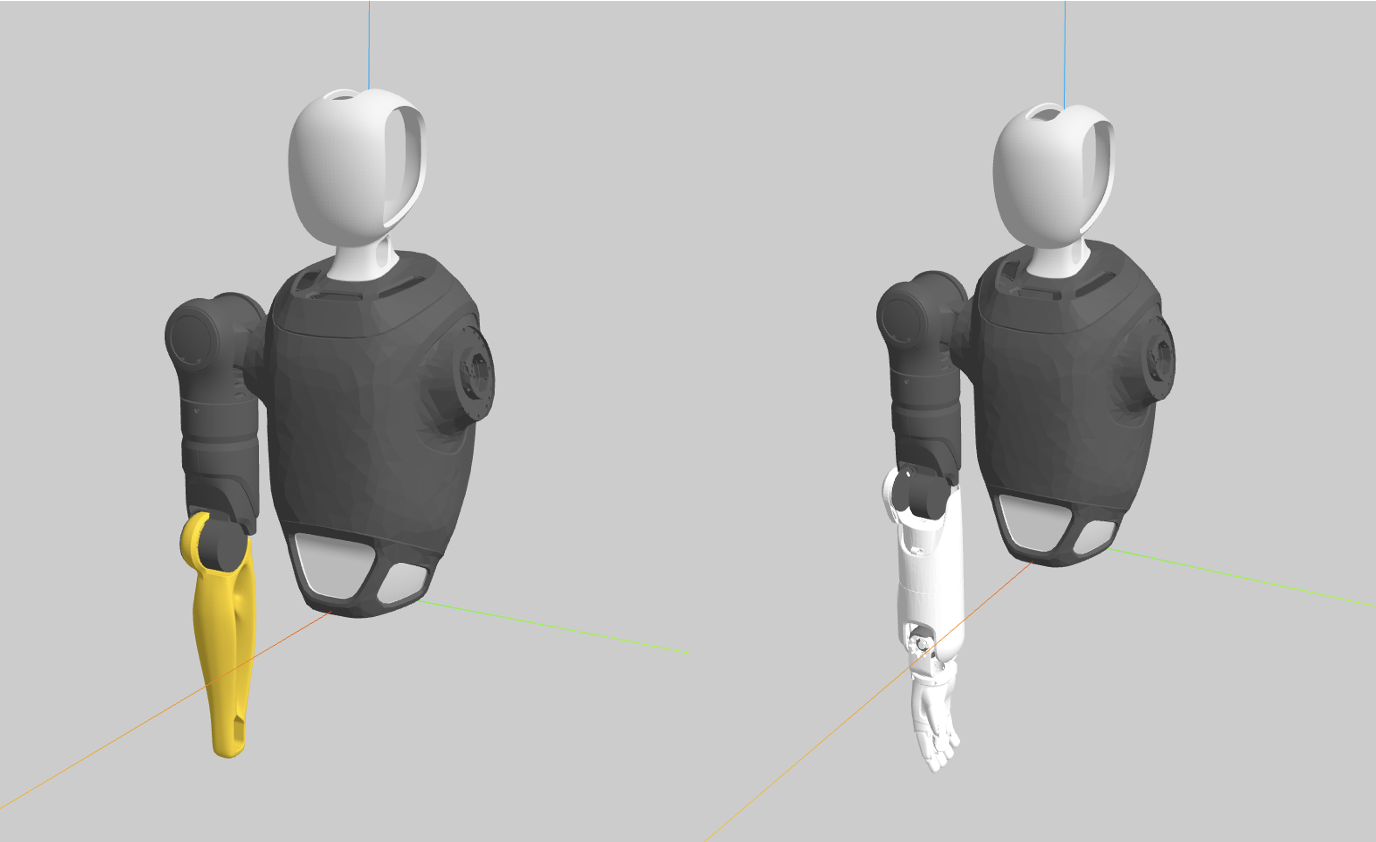}
    \caption{URDF modeling of the 4-DoF and 7-DoF arms.}
    \label{fig:arm_compare}
    \vspace{-0.3cm}
\end{figure}

The two arm configurations share the same mounting interface and upper-arm structure, enabling rapid replacement according to different application requirements. In this experiment, we quantitatively compare their reachable workspace and manipulation capability to analyze the contribution of the additional degrees of freedom. The corresponding arm configurations are illustrated in Fig.~\ref{fig:arm_compare}.

\subsection{Comparison Results and Discussion}
\subsubsection{Workspace and Manipulability}
\input{latex/tables/arm_comp}
The workspace is calculated for both arm configurations under identical base mounting conditions and joint limits. Since the forearm of the 7-DoF arm is longer than that of the 4-DoF arm, the maximum reachable workspace is naturally extended. Furthermore, the additional wrist degrees of freedom provide higher end-effector orientation flexibility.

For the 4-DoF arm, the Jacobian matrix cannot achieve full rank in the 6D Cartesian space, limiting independent control of the complete end-effector pose. Therefore, its dexterity is constrained in complex manipulation tasks requiring precise orientation adjustment. In contrast, the 7-DoF arm provides full 6D pose control with additional redundancy, enabling better singularity avoidance and joint configuration optimization, which improves manipulation flexibility and whole-body capability.

As for manipulability, the reduced 4-D manipulability is defined as
$w_{4}=\sqrt{\det(J^{\mathsf T}J)}$. For the 7-DoF arm, J5--J7 were fixed at zero and only the J1--J4 Jacobian columns were retained.

\subsubsection{End-effector Payload}
\input{latex/tables/arm_comp2}

\subsection{Summary}

The results verify the complementary advantages of the two arm configurations. 
The 7-DoF arm improves workspace and dexterity, while the 4-DoF arm provides 
higher payload efficiency and a lighter mechanical structure. The 
interchangeable design therefore enables flexible adaptation between dexterous 
manipulation and efficient operation requirements.

\section{Conclusions}

This supplementary experimental evaluation provides additional quantitative analysis and engineering assessments of the proposed X2-N transformable wheel-legged humanoid robot.

The energy evaluation results demonstrate the potential efficiency advantages of wheel-based locomotion, while the transformation experiments verify the repeatability and stability of the proposed morphology transition process under practical hardware conditions. The locking mechanism evaluation provides preliminary validation of the structural robustness during repeated transformation operations. The reinforcement learning ablation studies further reveal the contributions of different reward components to locomotion stability, trajectory tracking, and transformation performance. In addition, the comparison between different arm configurations analyzes the trade-off between manipulation capability and system complexity.

Overall, these experiments provide further evidence that the proposed transformable architecture enables flexible adaptation between efficient locomotion and multifunctional operation. Nevertheless, the current evaluations are conducted under limited experimental conditions, and more extensive experiments with larger trial numbers, longer-term operation, and diverse environments are still required to further improve and validate the robustness, durability, and reliability of the system.


%% file: ener_result.tex
\begin{table}[t]
\centering
\small

\caption{Power and energy evaluation.}
\label{tab:three_mode_locomotion_power_energy}

\setlength{\tabcolsep}{4pt}
\renewcommand{\arraystretch}{1.05}

\begin{tabular}{lccccc}
\toprule

Velocity &
Mode &
$P_{\mathrm{mech}}^{\mathrm{avg}}$ &
$P_{\mathrm{elec}}^{\mathrm{avg}}$ &
$\mathrm{CoT}_{\mathrm{mech}}$ &
$\mathrm{CoT}_{\mathrm{elec}}$
\\

&
&
(W) &
(W) &
&
\\

\midrule

\multirow{3}{*}{0.4 m/s}
& walk
& 38.44
& 133.29
& 0.498
& 1.723
\\

& wheel
& \textbf{10.73}
& \textbf{94.17}
& \textbf{0.077}
& \textbf{0.672}
\\

& hybrid
& --
& --
& --
& --
\\

\midrule

\multirow{3}{*}{0.6 m/s}
& walk
& 35.33
& 127.63
& 0.276
& 0.998
\\

& wheel
& \textbf{15.80}
& \textbf{99.15}
& \textbf{0.080}
& \textbf{0.501}
\\

& hybrid
& 61.65
& 110.94
& 0.366
& 0.659
\\

\midrule

\multirow{3}{*}{1.0 m/s}
& walk
& 94.09
& 173.12
& 0.895
& 1.650
\\

& wheel
& \textbf{32.28}
& \textbf{110.76}
& \textbf{0.094}
& \textbf{0.322}
\\

& hybrid
& 65.43
& 112.65
& 0.309
& 0.533
\\

\bottomrule

\end{tabular}

\end{table}

%% file: trans.tex
\begin{table}[H]
\centering
\small

\caption{Transformation performance evaluation.}
\label{tab:transform_result}

\renewcommand{\arraystretch}{1.3}
\setlength{\tabcolsep}{6pt}

\begin{tabular}{cccccc}
\toprule

Motion &
$T_{\mathrm{trans}}$ &
$e_{\mathrm{CoM}}$ &
$e_{\mathrm{ori}}$ &
\multicolumn{2}{c}{Success rate (\%)} \\

&
(s) &
(m) &
(deg) &
No-load &
Full-load \\

\midrule

Wheel to foot &
2.53 &
0.309 &
28.48 &
100 &
90.0 \\

Foot to wheel &
1.85 &
0.183 &
4.58 &
100 &
100 \\

\bottomrule

\end{tabular}

\end{table}

%% file: latex/tables/exp_overview.tex
\begin{table}[H]
\centering
\caption{Organization of the supplementary experiments.}
\label{tab:exp_overview}

\renewcommand{\arraystretch}{1.40}
\setlength{\tabcolsep}{8pt}

\begin{tabular}{p{1.2cm}p{10cm}p{0.8cm}}
\toprule
\textbf{Exp.} & \textbf{Title} & \textbf{Sec.}\\
\midrule

Exp. 1 & 
Energy efficiency evaluation under different locomotion modes. & 3\\

Exp. 2 &
Repeatability and stability evaluation of bidirectional transformation.  & 4\\

Exp. 3 &
Engineering validation of the locking mechanism.  & 5\\

Exp. 4 &
Ablation studies of the reinforcement learning reward components.  & 6\\

Exp. 5 &
Comparison evaluation of 4-DoF and 7-DoF interchangeable arms& 7\\

\bottomrule
\end{tabular}
\end{table}

%% file: latex/tables/robot_spec.tex
\begin{table}[H]
\centering
\caption{Robot specifications of the X2-N prototype.}
\label{tab:robot_spec}
\renewcommand{\arraystretch}{1.15}

\begin{tabular}{p{5.2cm}p{6.2cm}}
\toprule
\textbf{Item} & \textbf{Specification} \\
\midrule

Dimensions (H $\times$ L $\times$ W) &
$1.30 \times 0.47 \times 0.23$ m \\

Mass &
30.7 kg \\

Degrees of Freedom (DoF) &
21 (Foot-legged Mode) / 17 (Wheel-legged Mode) \\

Nominal Battery Voltage &
48 V \\

Wheel Diameter &
180 mm \\

Command Frequency &
50 Hz \\

Joint Control Frequency &
1000 Hz \\

\bottomrule
\end{tabular}

\end{table}

%% file: latex/tables/basic_envi_setup.tex

\begin{table}[H]
\centering
\small

\caption{Simulation and hardware experimental setups.}
\label{tab:exp_setup}

\renewcommand{\arraystretch}{1.25}
\setlength{\tabcolsep}{5pt}

\begin{tabular}{
    p{3.0cm}
    p{4.2cm}
    p{3.2cm}
}

\toprule

\textbf{Item} &
\textbf{Simulation} &
\textbf{Hardware}
\\

\midrule

Platform &
\makecell[l]{
Isaac Gym (RL Training)\\
MuJoCo (Sim-to-Sim Evaluation)
} &
\makecell[l]{
Indoor laboratory\\
($15\,\mathrm{m}\times7\,\mathrm{m}$)
}
\\

Ground surface &
Flat ground &
Flat indoor PVC floor
\\

Safety protection &
-- &
Overhead crane
\\

Computing platform &
\makecell[l]{
Desktop PC\\
(NVIDIA RTX 4090 GPU)
} &
Onboard computer
\\

\midrule

Operating system &
\multicolumn{2}{c}{
Ubuntu 22.04
}
\\

\bottomrule

\end{tabular}

\end{table}

%% file: latex/tables/basic_data.tex
\begin{table}[H]
\centering
\caption{Recorded robot states during the experiments.}
\label{tab:basic_data}
\renewcommand{\arraystretch}{1.15}

\begin{tabular}{p{3.0cm}p{6.5cm}cc}
\toprule
\textbf{Item} &
\textbf{Recorded Signals} &
\textbf{Sim.} &
\textbf{Hw.} \\
\midrule

Joint &
Position, velocity, torque &
\checkmark &
\checkmark \\

IMU &
Acceleration, angular velocity, orientation &
\checkmark &
\checkmark \\

Body (CoM) States &
Position, velocity, orientation &
\checkmark &
-- \\

Battery States &
Voltage, current &
-- &
\checkmark \\

Contact Forces &
Normal force, lateral force &
\checkmark &
-- \\

\bottomrule
\end{tabular}

\end{table}

%% file: latex/tables/ener_metric.tex

\begin{table}[H]
\centering

\caption{Evaluation metrics used throughout the energy experiments.}
\label{tab:ener_metric}

\renewcommand{\arraystretch}{0.95}
\setlength{\tabcolsep}{3pt}

\begin{tabular}{
>{\raggedright\arraybackslash}m{2.2cm}
>{\centering\arraybackslash}m{3.8cm}
>{\raggedright\arraybackslash}m{2.8cm}
}

\toprule

\textbf{Metric} &
\textbf{Formula (Unit)} &
\textbf{Notes}
\\

\midrule

Total Mechanical Power
&
$\displaystyle
P_{\mathrm{mech}}
=
\sum_{i=1}^{N}
|\tau_i\omega_i|
\;(\mathrm{W})
$
&
$\tau_i$: torque;
$\omega_i$: angular velocity;
$N$: actuators.
\\

\midrule

Electrical Power
&
$\displaystyle
P_{\mathrm{elec}}=VI
\;(\mathrm{W})
$
&
$V$: voltage;
$I$: BMS current.
\\

\midrule

Energy Consumption
&
$\displaystyle
E=\sum_{k=1}^{K}P_k\Delta t
\;(\mathrm{J})
$
&
$P_k$: sampled power;
$\Delta t$: interval.
\\

\midrule

Cost of Transport
&
$\displaystyle
CoT=\frac{E}{mgd}
$
&
$m$: mass;
$g$: gravity;
$d$: distance.
\\

\bottomrule

\end{tabular}

\end{table}

%% file: latex/tables/ener_setup.tex

\begin{table}[H]
\centering

\caption{Experimental settings for power and energy evaluation.}
\label{tab:ener_setup}

\renewcommand{\arraystretch}{1.1}
\setlength{\tabcolsep}{5pt}

\begin{tabular}{
>{\centering\arraybackslash}p{2.0cm}
>{\centering\arraybackslash}p{2.2cm}
>{\centering\arraybackslash}p{3.0cm}
>{\centering\arraybackslash}p{3.5cm}
}

\toprule

\textbf{Item} &
\textbf{Sub-item} &
\textbf{Simulation} &
\textbf{Hardware Experiment}
\\

\midrule

Still standing
&
-- 
&
10 s
&
5 mins
\\

\midrule

\multirow{3}{*}{Linear motion}
&
Travel distance
&
16 m straight path
&
\makecell[c]{8 m straight path\\
10 round trips\\
160 m in total}
\\

\cmidrule(lr){2-4}

&
Speed commands
&
0.5 m/s, 1 m/s, 1.5 m/s
&
0.4 m/s, 0.6 m/s, 1.0 m/s
\\

\cmidrule(lr){2-4}

&
Loco. Mode
&
\multicolumn{2}{c}{
Foot-legged, Wheel-legged, Wheel-legged*, Hybrid
}
\\

\bottomrule

\end{tabular}

\vspace{2pt}

\end{table}

%% file: latex/tables/ener_simu_result2.tex
\begin{table*}[t]
\centering
\scriptsize
\caption{Power and energy evaluation in simulation experiments.}
\label{tab:ener_simu_result2}
\setlength{\tabcolsep}{3.8pt}
\resizebox{0.86\textwidth}{!}{%
\begin{tabular}{llccccc}
\toprule
Motion & Velocity & Loco.\ Mode & \(P_{\mathrm{mech}}^{\mathrm{avg}}\) & \(P_{\mathrm{mech}}^{\max}\) & \(E_{\mathrm{mech},16\mathrm{m}}\) & \(\mathrm{CoT}_{\mathrm{mech}}\) \\
 & (\(\mathrm{m\,s^{-1}}\)) & & (\(\mathrm{W}\)) & (\(\mathrm{W}\)) & (\(\mathrm{J}\)) & \\
\midrule
Still standing & 0.0 & -- & 0.12\,$\pm$\,0.10 & 0.90 & -- & -- \\
\midrule
\multirow{12}{*}{Linear motion} & \multirow{4}{*}{0.5} & foot & 42.46\,$\pm$\,1.73 & 153.24 & 1923.1\,$\pm$\,77.7 & 0.390\,$\pm$\,0.016 \\
 &  & wheel & 18.21\,$\pm$\,0.22 & 19.78 & 402.3\,$\pm$\,31.6 & 0.082\,$\pm$\,0.006 \\
 &  & wheel$^{*}$ & \textbf{7.50\,$\pm$\,0.17} & \textbf{11.24} & \textbf{238.8\,$\pm$\,7.9} & \textbf{0.048\,$\pm$\,0.002} \\
 &  & hybrid & 42.76\,$\pm$\,4.24 & 113.18 & 975.0\,$\pm$\,89.6 & 0.198\,$\pm$\,0.018 \\
\cmidrule(lr){2-7}
 & \multirow{4}{*}{1.0} & foot & 113.00\,$\pm$\,3.56 & 595.06 & 1958.3\,$\pm$\,66.4 & 0.397\,$\pm$\,0.013 \\
 &  & wheel & 41.52\,$\pm$\,0.75 & 44.65 & 598.4\,$\pm$\,46.2 & 0.121\,$\pm$\,0.009 \\
 &  & wheel$^{*}$ & \textbf{33.31\,$\pm$\,1.07} & \textbf{37.56} & \textbf{490.7\,$\pm$\,38.2} & \textbf{0.099\,$\pm$\,0.008} \\
 &  & hybrid & 74.71\,$\pm$\,4.09 & 177.35 & 838.3\,$\pm$\,106.8 & 0.170\,$\pm$\,0.022 \\
\cmidrule(lr){2-7}
 & \multirow{4}{*}{1.5} & foot & -- & -- & -- & -- \\
 &  & wheel & 62.37\,$\pm$\,1.92 & 63.46 & 731.7\,$\pm$\,52.0 & 0.148\,$\pm$\,0.011 \\
 &  & wheel$^{*}$ & \textbf{38.33\,$\pm$\,2.93} & \textbf{43.90} & \textbf{535.8\,$\pm$\,56.5} & \textbf{0.109\,$\pm$\,0.011} \\
 &  & hybrid & 220.50\,$\pm$\,266.65 & 937.75 & 2710.6\,$\pm$\,1782.7 & 0.549\,$\pm$\,0.361 \\
\bottomrule
\end{tabular}%
}
\vspace{2pt}
\end{table*}

%% file: latex/tables/ener_hw_result2.tex

\begin{table*}[t]
\centering
\scriptsize
\caption{Power and energy evaluation in hardware experiments.}
\label{tab:locomotion_power_energy}
\setlength{\tabcolsep}{4.5pt}

\begin{tabular}{llccccccccc}
\toprule
Motion
& Velocity
& Loco. Mode
& $P_{\mathrm{mech}}^{\mathrm{avg}}$
& $P_{\mathrm{elec}}^{\mathrm{avg}}$
& $P_{\mathrm{mech}}^{\max}$
& $P_{\mathrm{elec}}^{\max}$
& $E_{\mathrm{elec},16\mathrm{m}}$
& $\mathrm{CoT}_{\mathrm{mech}}$
& $\mathrm{CoT}_{\mathrm{elec}}$
& $\eta$ \\
& (m/s) & & (W) & (W) & (W) & (W) & (J) & & & (\%) \\
\midrule

Still standing
& 0
& --
& $0.42 \pm 0.10$
& $74.04 \pm 0.86$
& 0.88
& 76.84
& 740.4
& --
& --
& 0.57 \\

\midrule

\multirow{12}{*}{Linear motion}

& \multirow{4}{*}{0.4}
& foot
& $38.44 \pm 2.61$
& $133.29 \pm 13.76$
& 409.40
& 1019.39
& $8504.6 \pm 1687.8$
& $0.498 \pm 0.093$
& $1.723 \pm 0.342$
& $28.9 \pm 2.9$ \\

&
& wheel
& $19.94 \pm 7.32$
& $111.33 \pm 17.90$
& 35.08
& 156.70
& $4479.8 \pm 720.4$
& $0.163 \pm 0.060$
& $0.908 \pm 0.146$
& $17.9 \pm 6.8$ \\

&
& wheel$^{*}$
& $\mathbf{10.73 \pm 0.53}$
& $\mathbf{94.17 \pm 2.90}$
& 42.81
& $\mathbf{108.01}$
& $\mathbf{3317.5 \pm 131.0}$
& $\mathbf{0.077 \pm 0.004}$
& $\mathbf{0.672 \pm 0.027}$
& $11.4 \pm 0.5$ \\

&
& hybrid
& --
& --
& --
& --
& --
& --
& --
& -- \\

\cmidrule(lr){2-11}

& \multirow{4}{*}{0.6}
& foot
& $35.33 \pm \text{--}$
& $127.63 \pm \text{--}$
& 155.87
& 195.06
& $4925.8 \pm \text{--}$
& $0.276 \pm \text{--}$
& $0.998 \pm \text{--}$
& $27.7 \pm \text{--}$ \\

&
& wheel
& $16.31 \pm 0.91$
& $117.26 \pm 2.96$
& 50.49
& 156.72
& $3073.8 \pm 98.8$
& $0.087 \pm 0.005$
& $0.623 \pm 0.020$
& $13.9 \pm 0.6$ \\

&
& wheel$^{*}$
& $\mathbf{15.80 \pm 1.03}$
& $\mathbf{99.15 \pm 4.18}$
& 57.88
& $\mathbf{121.22}$
& $\mathbf{2471.7 \pm 146.3}$
& $\mathbf{0.080 \pm 0.003}$
& $\mathbf{0.501 \pm 0.030}$
& $15.9 \pm 0.9$ \\

&
& hybrid
& $61.65 \pm 5.75$
& $110.94 \pm 18.83$
& 484.45
& 402.60
& $3254.7 \pm 582.5$
& $0.366 \pm 0.037$
& $0.659 \pm 0.118$
& $55.5 \pm 10.9$ \\

\cmidrule(lr){2-11}

& \multirow{4}{*}{1.0}
& foot
& $94.09 \pm 12.08$
& $173.12 \pm 18.10$
& 191.45
& 467.02
& $8141.9 \pm 2231.6$
& $0.895 \pm 0.244$
& $1.650 \pm 0.452$
& $54.2 \pm 6.7$ \\

&
& wheel
& $33.06 \pm 1.98$
& $122.76 \pm 3.63$
& 93.38
& 151.87
& $1740.9 \pm 63.6$
& $0.095 \pm 0.004$
& $0.353 \pm 0.013$
& $26.9 \pm 1.3$ \\

&
& wheel$^{*}$
& $\mathbf{32.28 \pm 1.35}$
& $\mathbf{110.76 \pm 2.67}$
& $\mathbf{91.21}$
& $\mathbf{139.91}$
& $\mathbf{1589.7 \pm 79.0}$
& $\mathbf{0.094 \pm 0.003}$
& $\mathbf{0.322 \pm 0.016}$
& $29.1 \pm 1.5$ \\

&
& hybrid
& $65.43 \pm 5.08$
& $112.65 \pm 11.22$
& 349.71
& 240.26
& $2629.3 \pm 315.1$
& $0.309 \pm 0.026$
& $0.533 \pm 0.064$
& $58.0 \pm 7.4$ \\

\bottomrule
\end{tabular}

\end{table*}

%% file: latex/tables/transform_metric.tex

\begin{table}[H]
\centering
\small

\caption{Evaluation metrics for bidirectional transformation experiments.}
\label{tab:transform_metric}

\renewcommand{\arraystretch}{1.3}
\setlength{\tabcolsep}{7pt}

\begin{tabular}{
>{\raggedright\arraybackslash}m{2.6cm}
>{\centering\arraybackslash}m{3.8cm}
>{\raggedright\arraybackslash}m{2.8cm}
}

\toprule

\textbf{Metric} &
\textbf{Formula (Unit)} &
\textbf{Notes}
\\

\midrule

Transformation Time
&
$T_{\mathrm{trans}}=t_e-t_s$
(s)
&
$t_s$: start; $t_e$: end.
\\

\midrule

Body CoM Displacement
&
$\displaystyle
e_{\mathrm{CoM}}
=
\left\|
p_{\mathrm{CoM}}^{e}
-
p_{\mathrm{CoM}}^{s}
\right\|_2
$
(m)
&
CoM displacement between start and end.
\\

\midrule

Body Orientation Deviation
&
$\displaystyle
e_{\mathrm{ori}}
=
\cos^{-1}
\left(
\frac{\mathrm{tr}(R_s^TR_e)-1}{2}
\right)
$
(rad)
&
$R_s$, $R_e$: initial and final orientations.
\\

\midrule

Transformation Energy
&
$\displaystyle
E_{\mathrm{trans}}
=
\sum_{k=s}^{e}P_k\Delta t
$
(J)
&
$P_k$: measured power; $\Delta t$: interval.
\\

\midrule

Success Rate
&
$\displaystyle
R_{\mathrm{success}}
=
\frac{N_s}{N_a}\times100\%
$
(\%)
&
$N_s$: successful trials; $N_a$: total trials.
\\

\bottomrule

\end{tabular}

\end{table}

%% file: latex/tables/transform_setup.tex
\begin{table}[H]
\centering
\small

\caption{Experimental settings for transformation repeatability evaluation.}
\label{tab:transformation_trials}

\renewcommand{\arraystretch}{1.0}
\setlength{\tabcolsep}{3pt}

\begin{tabular}{ccc}
\toprule

\textbf{Item} &
\textbf{No-load Transformation} &
\textbf{Full-body-load Transformation}
\\

\midrule

Number of cycles &
150 &
30
\\

\bottomrule

\end{tabular}

\end{table}

%% file: latex/tables/transform_result.tex

\begin{table}[H]
\centering
\small

\caption{Real-robot transformation metrics under full action execution.}
\label{tab:trans_results}

\renewcommand{\arraystretch}{1.0}
\setlength{\tabcolsep}{6pt}

\begin{tabular}{
>{\centering\arraybackslash}p{2cm}
>{\centering\arraybackslash}p{1.5cm}
>{\centering\arraybackslash}p{1.7cm}
>{\centering\arraybackslash}p{1.7cm}
>{\centering\arraybackslash}p{1.7cm}
>{\centering\arraybackslash}p{1.9cm}
>{\centering\arraybackslash}p{2.2cm}
}

\toprule

Motion &
$T_{\mathrm{trans}}$
(s) &
$e_{\mathrm{CoM}}$
(m) &
$e_{\mathrm{ori}}$
(deg) &
$e_{\mathrm{roll}}$
(deg) &
$e_{\mathrm{pitch}}$
(deg) &
$e_{\mathrm{yaw}}$
(deg)
\\

\midrule

Wheel to foot &
$2.53\pm0.25$ &
$0.309\pm0.076$ &
$28.48\pm10.64$ &
$0.680\pm0.696$ &
$-4.387\pm1.075$ &
$-27.909\pm11.292$
\\

Foot to wheel &
$1.85\pm0.50$ &
$0.183\pm0.083$ &
$4.58\pm1.39$ &
$0.416\pm0.450$ &
$3.952\pm1.489$ &
$-1.801\pm1.247$
\\

\bottomrule

\end{tabular}

\end{table}

%% file: latex/tables/transform_result2.tex

\begin{table}[H]
\centering
\small

\caption{Real-robot transformation power and energy evaluation.}
\label{tab:trans_results2}

\renewcommand{\arraystretch}{0.9}
\setlength{\tabcolsep}{2.5pt}

\begin{tabular}{
>{\centering\arraybackslash}p{1.8cm}
>{\centering\arraybackslash}p{2.5cm}
>{\centering\arraybackslash}p{1.8cm}
>{\centering\arraybackslash}p{2.5cm}
>{\centering\arraybackslash}p{2.5cm}
}

\toprule

Motion &
$E_{\mathrm{mech}}$ (J) &
$P_{\mathrm{mech}}$ (W) &
$E_{\mathrm{elec}}$ (J) &
$P_{\mathrm{elec}}$ (W)
\\

\midrule

Wheel-to-foot &
$219.90\pm19.72$ &
$87.14\pm6.84$ &
$539.09\pm110.38$ &
$214.67\pm47.55$
\\

Foot-to-wheel &
$128.24\pm16.70$ &
$72.62\pm16.23$ &
$210.21\pm101.16$ &
$115.41\pm45.72$
\\

\bottomrule

\end{tabular}

\end{table}

%% file: latex/tables/trans_succ.tex

\begin{table}[H]
\centering
\small

\caption{Success rate evaluation of bidirectional transformation under 
different loading conditions.}
\label{tab:trans_success_rate}

\renewcommand{\arraystretch}{1.1}
\setlength{\tabcolsep}{6pt}

\begin{tabular}{ccccc}
\toprule

Condition &
Motion &
$N_s$ &
$N_a$ &
$R_{\mathrm{success}}$ (\%)
\\

\midrule

\multirow{2}{*}{No-load}
& Wheel to foot
& 150
& 150
& 100
\\

& Foot to wheel
& 150
& 150
& 100
\\

\midrule

\multirow{2}{*}{Full-body-load}
& Wheel to foot
& 27
& 30
& 90
\\

& Foot to wheel
& 30
& 30
& 100
\\

\bottomrule

\end{tabular}

\end{table}

%% file: latex/tables/locking_metric.tex

\begin{table}[H]
\centering
\small

\caption{Evaluation metrics for locking mechanism validation.}
\label{tab:locking_metric}

\renewcommand{\arraystretch}{1.15}
\setlength{\tabcolsep}{4pt}

\begin{tabular}{
>{\raggedright\arraybackslash}m{2.8cm}
>{\centering\arraybackslash}m{2.8cm}
>{\raggedright\arraybackslash}m{3.6cm}
}

\toprule

\textbf{Metric} &
\textbf{Formula (Unit)} &
\textbf{Notes}
\\

\midrule

Maximum Impact Load
&
$F_{\mathrm{impact,max}}$ (N)
&
Maximum impact force during locking engagement.
\\

\midrule

Maximum Impact Torque
&
$\tau_{\mathrm{impact,max}}$ (Nm)
&
Maximum joint torque during locking engagement.
\\

\midrule

Locking Backlash
&
$\Delta \theta$ (arcmin)
&
Displacement caused by mechanical clearance after locking.
\\

\midrule

Wear Degradation
&
$\Delta D$ (mm)
&
Clearance increase after repeated locking cycles.
\\

\midrule

Success Rate
&
$\displaystyle
R_{\mathrm{success}}
=
\frac{N_s}{N_a}\times100\%
$
(\%)
&
$N_s$: successful operations;
$N_a$: total attempts.
\\

\bottomrule

\end{tabular}

\end{table}

%% file: latex/tables/locking_setup.tex

\begin{table}[H]
\centering
\small

\caption{Validation setups for the locking mechanism.}
\label{tab:locking_validation}

\renewcommand{\arraystretch}{1.1}
\setlength{\tabcolsep}{4pt}

\begin{tabular}{
>{\centering\arraybackslash}p{2.0cm}
>{\centering\arraybackslash}p{3.0cm}
>{\centering\arraybackslash}p{2.3cm}
>{\centering\arraybackslash}p{2.5cm}
}

\toprule

\textbf{Item} &
\textbf{Simulation} &
\multicolumn{2}{c}{
\textbf{Hardware}
}
\\

\midrule

Condition &
SolidWorks Analysis &
No Load &
Full Robot Load
\\

Evaluation &
\makecell[c]{Finite Element\\Analysis}
&
\makecell[c]{Ankle-level\\Transformation}
&
\makecell[c]{Whole Robot\\Transformation}
\\

Number of Trials &
-- &
150 &
30
\\

\bottomrule

\end{tabular}

\end{table}

%% file: latex/tables/lock_load.tex

\begin{table}[H]
\centering
\small

\caption{Maximum loading conditions of the locking mechanism.}
\label{tab:locking_validation}

\renewcommand{\arraystretch}{1.1}
\setlength{\tabcolsep}{5pt}

\begin{tabular}{cccc}

\toprule

\textbf{Item} & \textbf{Max. Impact Normal Force} &
\textbf{Max. Impact Lateral Force} &
\textbf{Max. Rotation Torque}
\\

\midrule

Condition & 
\makecell[c]{Hybrid mode\\stepping}
&
\makecell[c]{High-speed\\turning}
&
\makecell[c]{High-speed\\rolling acceleration}
\\

\midrule
Value &
900 N
&
100 N
&
20 Nm
\\

\bottomrule

\end{tabular}

\end{table}

%% file: latex/tables/exp4_sub_overview.tex

\begin{table}[H]
\centering
\small

\caption{Organization of the supplementary experiments.}
\label{tab:ablation_exp_list}

\renewcommand{\arraystretch}{1.1}
\setlength{\tabcolsep}{4pt}

\begin{tabular}{
>{\centering\arraybackslash}p{1.5cm}
>{\centering\arraybackslash}p{3cm}
>{\centering\arraybackslash}p{6cm}
>{\centering\arraybackslash}p{1.8cm}
}

\toprule

\textbf{Exp.} &
\textbf{RL Agent Type} &
\textbf{Title} &
\textbf{Key Reward}
\\

\midrule

Exp. 4.1 &
Wheel-legged Agent &
Centrifugal Compensation Reward Ablation &
$R_{\mathrm{cen}}$
\\

Exp. 4.2 &
Foot-legged Agent &
Foot-legged Walking Mimic Reward Ablation &
$R_{\mathrm{mimic}}$
\\

Exp. 4.3 &
Transformation Agent &
Transformation Balance Reward Ablation &
$R_{\mathrm{balance}}$
\\

\bottomrule

\end{tabular}

\end{table}

%% file: latex/tables/anti_cen_metric.tex

\begin{table}[H]
\centering

\caption{Evaluation metrics for anti-centrifugal force reward ablation.}
\label{tab:anti_cen_metrics}

\renewcommand{\arraystretch}{1.1}
\setlength{\tabcolsep}{4pt}

\begin{tabular}{
p{2.5cm}
p{4.0cm}
p{5.5cm}
}

\toprule

\textbf{Metric} &
\textbf{Formula (Unit)} &
\textbf{Notes}
\\

\midrule

Roll Tilt Angle &
$\theta_{\mathrm{roll}}=|\phi|$
(deg)
&
$\phi$: measured body roll angle; represents lateral inclination.
\\

\midrule

Velocity Tracking Error &
$\displaystyle
e_v=
\frac{1}{T}
\int_0^T
|v-v_d|dt
$
(m/s)
&
$v$: measured velocity; 
$v_d$: commanded velocity.
\\

\midrule

Turning Radius Error &
$\displaystyle
e_r=|r-r_d|
$
(m)
&
$r$: measured turning radius; 
$r_d$: desired turning radius.
\\

\midrule

Centripetal Force &
$\displaystyle
F_c=\frac{mv^2}{r}
$
(N)
&
$m$: robot mass; 
$v$: velocity; 
$r$: turning radius.
\\

\midrule

Centripetal Force Error &
$\displaystyle
e_c=
|F_c^a-F_c^d|
$
(N)
&
$F_c^a$: actual centripetal force; 
$F_c^d$: required force from commanded velocity and reference radius.
\\
\midrule
Turning Radius Error &
$\displaystyle
e_r=
\frac{|r-r_d|}{r_d}
$
(Unitless)
&
$r$: measured turning radius;
$r_d$: desired turning radius.
\\

\bottomrule

\end{tabular}

\end{table}

%% file: latex/tables/anti_cen_setup.tex
\begin{table}[H]
\centering
\caption{Simulation settings for circular motion evaluation.}
\label{tab:circular_motion_setup}

\renewcommand{\arraystretch}{1.3}
\setlength{\tabcolsep}{10pt}

\begin{tabular}{
>{\raggedright\arraybackslash}p{3.0cm}
>{\raggedright\arraybackslash}p{7.0cm}
}
\toprule

\textbf{Item} &
\textbf{Simulation} \\

\midrule

Motion &
Circular motion with constant diameter (4 m) \\

Linear Velocity &
0.5 m/s, 1 m/s, 1.2 m/s \\

Angular Velocity &
0.25 rad/s, 0.5 rad/s, 0.5 rad/s \\

\bottomrule

\end{tabular}

\end{table}

%% file: latex/tables/anti_cen_result.tex

\begin{table}[H]
\centering

\caption{Ablation results of the anti-centrifugal reward during circular motion.}
\label{tab:anti_cen_ablation}

\setlength{\tabcolsep}{3pt}
\renewcommand{\arraystretch}{1.0}

\begin{tabular}{cccccccccc}
\toprule

$v_x$ &
$|\omega_z|$ &
$r_d$ &
Method &
$r$ &
$e_r$ &
$|\phi|$ &
$e_v$ &
$F_c$ &
$e_c$
\\

(m/s) &
(rad/s) &
(m) &
&
(m) &
&
(deg) &
(m/s) &
(N) &
(N)
\\

\midrule

\multirow{2}{*}{0.5}
&
\multirow{2}{*}{0.25}
&
\multirow{2}{*}{2.0}
&
w/o reward
& 
1.693
&
0.1535
&
1.931
&
0.0113
&
4.437
&
\textbf{0.508}
\\

&
&
&
w/ reward
&
\textbf{1.643}
&
\textbf{0.1785}
&
\textbf{2.210}
&
\textbf{0.0038}
&
\textbf{4.709}
&
0.786
\\

\midrule

\multirow{2}{*}{1.0}
&
\multirow{2}{*}{0.50}
&
\multirow{2}{*}{2.0}
&
w/o reward
&
2.567
&
0.2835
&
0.923
&
0.0522
&
10.992
&
4.719
\\

&
&
&
w/ reward
&
\textbf{2.055}
&
\textbf{0.0275}
&
\textbf{5.462}
&
\textbf{0.0386}
&
\textbf{16.420}
&
\textbf{0.785}
\\

\midrule

\multirow{2}{*}{1.2}
&
\multirow{2}{*}{0.50}
&
\multirow{2}{*}{2.4}
&
w/o reward
&
3.570
&
0.4875
&
0.329
&
0.0731
&
11.204
&
7.681
\\

&
&
&
w/ reward
&
\textbf{2.321}
&
\textbf{0.0329}
&
\textbf{5.863}
&
\textbf{0.0419}
&
\textbf{18.054}
&
\textbf{0.696}
\\

\bottomrule

\end{tabular}

\end{table}

%% file: latex/tables/mimic_setup.tex
\begin{table}[H]
\centering
\caption{Experimental settings for mimic reward evaluation.}
\label{tab:mimic_setup}

\renewcommand{\arraystretch}{1.3}
\setlength{\tabcolsep}{10pt}

\begin{tabular}{lc}
\toprule

\textbf{Item} &
\textbf{Configuration} \\

\midrule

Motion &
Straight-line walking \\

Velocity &
0.4 m/s, 0.7 m/s, 1.0 m/s \\

Reference Trajectory & Predefined Joint Trajectory\\

\bottomrule

\end{tabular}

\end{table}

%% file: latex/tables/mimic_metric.tex

\begin{table}[H]
\centering
\small

\caption{Evaluation metrics for mimic reward ablation.}
\label{tab:mimic_metrics}

\renewcommand{\arraystretch}{1.0}
\setlength{\tabcolsep}{4pt}

\begin{tabular}{
p{3.0cm}
p{3.8cm}
p{5.6cm}
}

\toprule

\textbf{Metric} &
\textbf{Formula (Unit)} &
\textbf{Notes}
\\

\midrule

Joint Position Error &
$\displaystyle
e_{q_i}=|q_i-q_i^{ref}|
$
(deg)
&
$q_i$: measured joint position;  
$q_i^{ref}$: reference joint position.
\\

\bottomrule

\end{tabular}

\end{table}

%% file: latex/tables/mimic_result.tex

\begin{table}[H]
\centering
\small

\caption{Mimic reward ablation results.}
\label{tab:mimic_reward_ablation_test6}

\renewcommand{\arraystretch}{1.05}
\setlength{\tabcolsep}{5pt}

\begin{tabular}{cccc}
\toprule

Velocity &
Method &
$e_{q_i}$ (deg) &
$\Delta e_q$ (\%)
\\

\midrule

\multirow{2}{*}{0.4 m/s}
& w/o $R_{\mathrm{mimic}}$
& 13.23
& --
\\

& w/ $R_{\mathrm{mimic}}$
& 4.22
& +68.1\%
\\

\midrule

\multirow{2}{*}{0.7 m/s}
& w/o $R_{\mathrm{mimic}}$
& 20.50
& --
\\

& w/ $R_{\mathrm{mimic}}$
& 6.23
& +69.6\%
\\

\midrule

\multirow{2}{*}{1.0 m/s}
& w/o $R_{\mathrm{mimic}}$
& 25.01
& --
\\

& w/ $R_{\mathrm{mimic}}$
& 8.27
& +66.9\%
\\

\bottomrule

\end{tabular}

\end{table}

%% file: latex/tables/balance_setup.tex
\begin{table}[H]
\centering
\caption{Experimental settings for balance reward evaluation.}
\label{tab:balance_setup}

\renewcommand{\arraystretch}{1.3}
\setlength{\tabcolsep}{10pt}

\begin{tabular}{lc}
\toprule

\textbf{Item} &
\textbf{Configuration} \\

\midrule

Motion &
In-place transformation \\

Ground friction &
0.3, 0.5, 0.7, 1.0 \\

\bottomrule

\end{tabular}

\end{table}

%% file: latex/tables/balance_metric.tex
\begin{table}[H]
\centering
\small

\caption{Metrics in the balance reward ablation tests.}
\label{tab:balance_metric_definitions}

\renewcommand{\arraystretch}{1.1}
\setlength{\tabcolsep}{4pt}

\begin{tabular}{
p{2.8cm}
p{3.5cm}
p{4.6cm}
}
\toprule

\textbf{Metric} &
\textbf{Formula (Unit)} &
\textbf{Notes} \\

\midrule

Transformation Time &
$T_{\mathrm{trans}}
=
t_{\mathrm{end}}-t_{\mathrm{start}}$
(s) &
$t_{\mathrm{start}}$: start time;
$t_{\mathrm{end}}$: first stable state maintained for 0.5 s.
\\

\midrule

Maximum Tilt Angle &
$\displaystyle
\theta_{\max}
=
\max_t
\|
\boldsymbol{\theta}(t)
\|_2$
(deg) &
Maximum roll--pitch inclination during transformation.
\\

\midrule

CoM Position Error &
$\displaystyle
e_{\mathrm{CoM}}
=
\|
\mathbf{p}_{\mathrm{CoM}}^{e}
-
\mathbf{p}_{\mathrm{CoM}}^{s}
\|_2$
(m) &
CoM displacement between transformation start and end.
\\

\bottomrule

\end{tabular}

\end{table}

%% file: latex/tables/balance_result.tex
\begin{table}[H]
\centering
\small

\caption{Ablation results of the balance reward under different ground friction coefficients.}
\label{tab:balance_reward_ablation}

\setlength{\tabcolsep}{4pt}
\renewcommand{\arraystretch}{1.05}

\begin{tabular}{clcccc}
\toprule

Friction &
Method &
$T_{\mathrm{trans}}$
&
$\theta_{\max}$ (deg)
&
$e_{\mathrm{CoM}}$
&
$R_{\mathrm{success}}$
\\

\midrule

\multirow{2}{*}{0.3}
& w/o $R_{\mathrm{balance}}$
& 2.324
& \textbf{12.55}
& 0.447
& 100\%
\\

& w/ $R_{\mathrm{balance}}$
& \textbf{1.349}
& 18.79
& \textbf{0.232}
& 100\%
\\

\midrule

\multirow{2}{*}{0.5}
& w/o $R_{\mathrm{balance}}$
& 2.392
& \textbf{11.57}
& 0.403
& 100\%
\\

& w/ $R_{\mathrm{balance}}$
& \textbf{1.579}
& 20.34
& \textbf{0.138}
& 100\%
\\

\midrule

\multirow{2}{*}{0.7}
& w/o $R_{\mathrm{balance}}$
& 2.444
& \textbf{11.69}
& 0.438
& 100\%
\\

& w/ $R_{\mathrm{balance}}$
& \textbf{2.026}
& 20.46
& \textbf{0.154}
& 100\%
\\

\midrule

\multirow{2}{*}{1.0}
& w/o $R_{\mathrm{balance}}$
& 2.440
& \textbf{11.74}
& 0.446
& 100\%
\\

& w/ $R_{\mathrm{balance}}$
& \textbf{1.430}
& 20.57
& \textbf{0.129}
& 100\%
\\

\bottomrule

\end{tabular}

\end{table}

%% file: latex/tables/arm_comp.tex
\begin{table}[H]
    \centering
    \caption{Workspace comparison between the 4-DoF and 7-DoF
    interchangeable arm configurations.}
    \label{tab:workspace_comparison}
    \begin{tabular}{lccc}
        \toprule
        \textbf{Workspace metric} &
        \textbf{4-DoF} &
        \textbf{7-DoF} &
        \textbf{Improvement} \\
        \midrule
        X-axis span
            & 0.9589 m
            & \textbf{1.2176 m}
            & $+26.98\%$ \\

        Y-axis span
            & 0.8580 m
            & \textbf{1.1063 m}
            & $+28.94\%$ \\

        Z-axis span
            & 0.9588 m
            & \textbf{1.2186 m}
            & $+27.10\%$ \\

        Maximum shoulder-referenced reach
            & 0.5465 m
            & \textbf{0.6785 m}
            & $+24.15\%$ \\

        Workspace volume
            & 0.28990 m$^{3}$
            & \textbf{0.64828 m$^{3}$}
            & $+123.62\%$ \\

        Mean translational manipulability
            & 0.010580 m$^{3}$
            & \textbf{0.022035 m$^{3}$}
            & $+108.27\%$ \\

        Mean reduced 4-D manipulability
            & 0.166356
            & \textbf{0.177072}
            & $+6.44\%$ \\
        \bottomrule
    \end{tabular}
\end{table}

%% file: latex/tables/arm_comp2.tex
\begin{table}[H]
    \centering
    \caption{Comparison of arm self-weight and static payload capacity.}
    \label{tab:arm_payload_comparison}

    \small
    \setlength{\tabcolsep}{4pt}
    \renewcommand{\arraystretch}{1.12}

    \begin{tabularx}{0.8\columnwidth}{
        >{\raggedright\arraybackslash}X
        >{\centering\arraybackslash}p{0.2\columnwidth}
        >{\centering\arraybackslash}p{0.2\columnwidth}
        >{\centering\arraybackslash}p{0.1\columnwidth}
    }
        \toprule
        \textbf{Metric} &
        \textbf{X2-N 4-DoF} &
        \textbf{X2-T2 7-DoF} &
        \textbf{Change} \\
        \midrule

        Arm self-weight
            & 1.739 kg
            & \textbf{2.686 kg}
            & $+54.5\%$ \\

        Worst-case static payload
            & \textbf{2.844 kg}
            & 1.255 kg
            & $-55.9\%$ \\

        Worst-case pose
            & Horizontal side reach
            & Forward bent-elbow reach
            & -- \\

        Payload at max.\ horizontal reach
            & \textbf{2.999 kg}
            & 1.422 kg
            & $-52.6\%$ \\
        \bottomrule
    \end{tabularx}
\end{table}